\documentclass{article}
\usepackage[preprint]{colm2026_conference}

\usepackage{microtype}
\usepackage{hyperref}
\usepackage{url}
\usepackage{booktabs}

\usepackage{lineno}
\usepackage{graphicx}
\usepackage{amsmath}
\usepackage{wrapfig}
\usepackage{xspace}
\usepackage[T1]{fontenc}
\usepackage[utf8]{inputenc}
\usepackage{booktabs}
\usepackage{tabularx}
\usepackage{makecell}
\usepackage{enumitem}

\definecolor{darkblue}{rgb}{0, 0, 0.5}
\hypersetup{colorlinks=true, citecolor=darkblue, linkcolor=darkblue, urlcolor=darkblue}

\def \method{\textsc{CURE}\xspace}
\usepackage{listings}
\usepackage{xcolor}

\title{Think Through Uncertainty: Improving Long-Form Generation Factuality via Reasoning Calibration}

\author{Xin Liu \& Lu Wang \\
Computer Science and Engineering\\
University of Michigan\\
Ann Arbor, MI, USA \\
\texttt{\{liuxincs, wangluxy\}@umich.edu} \\
}

\begin{document}

\ifcolmsubmission
\linenumbers
\fi

\maketitle

\begin{abstract}
Large language models (LLMs) often hallucinate in long-form generation. Existing approaches mainly improve factuality through post-hoc revision or reinforcement learning (RL) with correctness-based rewards, but these methods do not teach models to produce calibrated confidence estimates. 
As a result, models may still state incorrect claims confidently in their responses.
Recent advances in reasoning have significantly improved LLM performance, and have been leveraged to estimate confidence by incorporating calibration into RL objectives. However, existing approaches remain limited to a single scalar confidence for the entire response, which is insufficient for long-form generation where uncertainty varies across individual claims.
To mitigate this problem, we propose \textbf{\method}, a framework that improves long-form factuality by teaching LLMs to reason about uncertainty at the claim level. 
We first introduce a \emph{claim-aware reasoning protocol}, which structures outputs into atomic claims paired with explicit confidence estimates. 
We then develop a \textit{multi-stage training pipeline} that aligns model confidence with claims' correctness and then optimizes on factuality. The resulting calibrated confidence further enables selective prediction, allowing the model to abstain from uncertain claims at inference time.
Experiments on four long-form factuality benchmarks show that \method consistently improves factual accuracy over competitive supervised and RL baselines, while maintaining factual recall. In particular, it improves claim-level factual accuracy by up to 39.9\% on Biography generation.
These gains are accompanied by improved calibration, as reflected by a 16.0\% increase in AUROC on FactBench.\footnote{Additional details about the project are availble on its Huggingface page \url{https://huggingface.co/spaces/launch/cure}}
\end{abstract}

\section{Introduction}

Large language models (LLMs) have demonstrated impressive capabilities in generating long-form responses to open-ended queries \citep{DBLP:journals/corr/abs-2412-16720, DBLP:journals/corr/abs-2501-12948, DBLP:journals/corr/abs-2505-09388}. However, they still suffer from factual errors, especially in long-form generation, where mistakes can compound across multiple claims and become difficult for users to verify \citep{zhang2025siren}.
A fundamental challenge is that models lack awareness of their own uncertainty: they state incorrect claims with high confidence, which undermines the factual reliability of their outputs. 
Accurate uncertainty estimation, or calibration, is therefore critical, as it enables models to distinguish reliable from unreliable claims, improving factuality while providing transparency to the user.

Existing factuality improvement methods have mainly relied on post-hoc revision, where the model response is verified and corrected after generation \citep{dhuliawala-etal-2024-chain, chern2023factool}. While effective, these methods introduce additional inference cost and do not improve the model's intrinsic uncertainty awareness.
More recently, training-time approaches optimize factual correctness via Reinforcement Learning (RL), either through step-wise verification to guide token-level updates \citep{li2025reasoning}, or via composite rewards over precision, relevance, and response quality \citep{DBLP:journals/corr/abs-2508-05618}.
However, these methods optimize solely for correctness and ignore model's intrinsic uncertainty. As a result, they fail to distinguish reliable claims from speculative ones, allowing overconfident errors to persist in response.

Meanwhile, recent advances in reasoning have enabled LLMs to generate chain of thoughts, which provide a natural interface for uncertainty expression. 
Building on this, recent work incorporates calibration into RL objectives, training models to estimate confidence alongside their reasoning~\citep{damani2025beyond}. 
However, these approaches are primarily designed for short-form settings, where each input is associated with a single scalar confidence score. 
In contrast, long-form generation involves multiple claims with varying uncertainty, which cannot be captured by a global confidence estimate. 
As a result, existing RL-based calibration methods remain inadequate for improving the factuality of long-form generation.

\begin{wrapfigure}{t}{0.48\textwidth}
    \centering
    \includegraphics[width=0.47\textwidth]{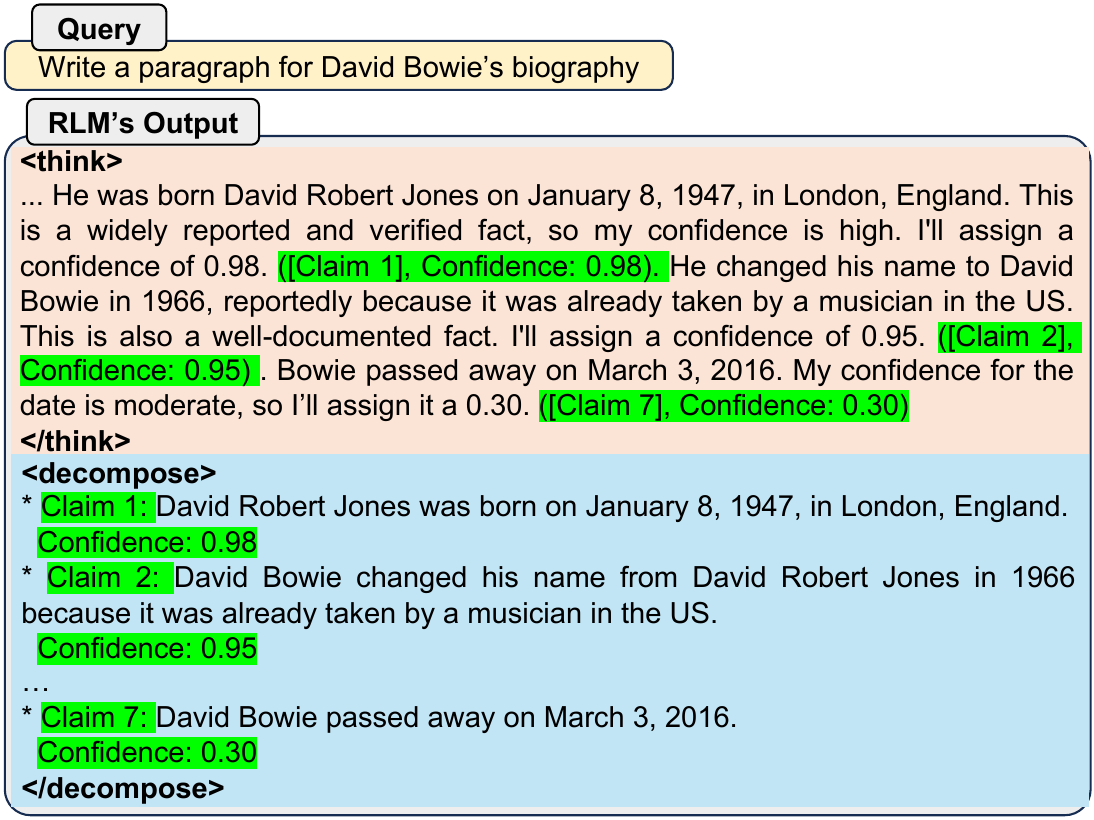}
    \caption{Illustration of the claim-aware reasoning protocol.The model performs structured reasoning with explicit uncertainty, then decomposes the output into atomic claims with calibrated confidence. These confidences enable selective prediction (Figure~\ref{fig:selective}).}
    \label{fig:format}
\end{wrapfigure}

Building on this observation, we propose \textbf{\method}, a \textbf{C}laim-level \textbf{U}ncertainty-aware \textbf{RE}soning framework that improves long-form factuality by enabling fine-grained, claim-level uncertainty modeling.
At the core is the \emph{claim-aware reasoning protocol} (Figure~\ref{fig:format}), where the model decomposes its response into atomic claims with explicit confidence estimates. For instance, when generating a biography, the model notes an unverified date and assigns the claim with a low confidence of 0.3. 
Next, using this protocol, we develop a \textit{multi-stage training pipeline} that disentangles calibration from factual optimization. We start with inducing a feasible reasoning space via supervised fine-tuning and a warm-up Group Relative Policy Optimization (GRPO) \citep{shao2024deepseekmath} stage. We then apply Direct Preference Optimization (DPO) \citep{rafailov2023direct} to align model-predicted confidence with empirical correctness, followed by GRPO with token-masked rewards to optimize claim-level factuality without disrupting calibration. At inference, calibrated confidences enable selective prediction, allowing the model to abstain from uncertain claims.

We evaluate our approach on four
long-form factuality benchmarks: FactBench~\citep{bayat2025factbench}, LongFact~\citep{wei2024long}, Biography~\citep{min2023factscore}, and FactRBench~\citep{liu-etal-2025-verifact}.
Across all datasets, \method consistently achieves better factual accuracy and strong calibration quality than comparisons. 
In particular, on FactBench, \method achieves a 9.4\% relative improvement in claim-level factual accuracy over a competitive RL baseline L2RF \citep{DBLP:journals/corr/abs-2508-05618},
while also achieving the highest AUROC (0.667).
Similar relative gains in factual accuracy are observed on LongFact (13.6\%) and Biography (39.9\%), further demonstrating the effectiveness of our approach in improving factuality.
On FactRBench, which provides reference claim sets for recall measurement,
\method maintains competitive factual recall compared to the base model, indicating that improved calibration and accuracy do not come at the cost of substantially reduced coverage.
These results highlight the critical advantage of modeling uncertainty at the claim level. By explicitly disentangling calibration from factual optimization, our framework effectively mitigates the overconfidence typically induced by standard RL, leading to more accurate and robust reasoning. 
Furthermore, the resulting fine-grained confidence estimates provide a reliable signal for distinguishing correct from incorrect claims.
Ultimately, this ensures that the generated responses are both highly controllable and genuinely trustworthy for end users.

\section{Related Work}

\paragraph{Long-form Factuality in LLMs}
Prior work on long-form factuality can be grouped into three paradigms. 
Post-hoc revision methods verify and refine model outputs after generation, either by decomposing responses into verification questions and revising accordingly~\citep{dhuliawala-etal-2024-chain}, or by detecting errors with external tools and performing targeted correction~\citep{chern2023factool}. 
A second line of work incorporates factuality signals into RL training, using either step-wise verification to guide token-level updates~\citep{li2025reasoning} or composite rewards that jointly optimize factual precision, response detail, and relevance~\citep{DBLP:journals/corr/abs-2508-05618}. 
A third line focuses on fine-grained preference alignment, where external evaluators provide sentence- or claim-level supervision to construct higher-quality preference data~\citep{huang2024factalign, gu2025mask, xie-etal-2025-improving}. 
Despite their effectiveness, these approaches optimize only for \emph{correctness} without modeling uncertainty, making it difficult to distinguish reliable claims from speculative ones.

\paragraph{LLM Calibration}
Another line of work studies how to estimate and calibrate model uncertainty. 
Existing approaches include prompting models to output scalar confidence scores~\citep{DBLP:conf/emnlp/TianMZSRYFM23}, estimating uncertainty via consistency across multiple sampled outputs~\citep{DBLP:conf/iclr/KuhnGF23,DBLP:journals/corr/abs-2306-13063}, and deriving confidence directly from token probabilities~\citep{DBLP:journals/corr/abs-2310-19208}. 
More recent work incorporates calibration into RL by augmenting correctness rewards with proper scoring rules such as the Brier score~\citep{damani2025beyond}. 
However, these methods are primarily designed for short-form settings with a single confidence per response, or operate at a coarse response level that cannot capture claim-level uncertainty in long-form generation.

Our work bridges these two directions by enabling claim-level calibration for long-form generation. 
We train the model to express uncertainty at the atomic claim level through structured reasoning, and decouple calibration from factuality optimization to avoid the interference observed in joint training.

\section{\method: Claim-level Uncertainty-aware Resoning for Factual Long-Form Generation}
\subsection{Problem Formulation}
Given an information-seeking query $x$, the model generates a factual response $y$. Such long-form responses typically contain multiple factual statements, which we represent as a set of atomic claims
$y=\{c_{1},c_{2},...,c_{N}\}$,
where each claim $c_{i}$ is a semantically coherent and independently verifiable content unit (e.g., "X was born in Y" as illustrated in Figure~\ref{fig:format}).

For each claim $c_{i}$, the model produces a confidence estimate $p_{i}$ between 0 and 1,
reflecting its internal belief in the correctness of that claim. Let $z_{i}\in\{0,1\}$ denote the ground-truth correctness label of $c_{i}$ evaluated against external knowledge. 
Our goal is to improve long-form factuality by generating claims that are both factually correct and comprehensively address the query.
To achieve this, the model must distinguish correct claims from speculative ones, which requires calibrated claim-level confidence estimates.

To this end, we introduce a claim-aware reasoning protocol (\S\ref{sec:protocol}) that structures outputs into atomic claims with explicit confidence, and a multi-stage training pipeline (\S\ref{sec:multi-stage}) that induces a feasible reasoning space, aligns confidence with correctness, and optimizes factuality.  
We further enable selective prediction at inference time (\S\ref{sec:selective_prediction}) based on the calibrated claim-level confidence.

\subsection{claim-aware reasoning protocol}
\label{sec:protocol}
To operationalize claim-level modeling, we introduce a \emph{claim-aware reasoning protocol} that structures generation and makes uncertainty explicit. 
As illustrated in Figure~\ref{fig:format}, given an input query $x$, the model first enters a \texttt{<think>} stage, where it constructs a reasoning trajectory, identifies candidate claims, and explicitly expresses uncertainty. 
The reasoning is then transformed into a structured set of atomic claims via a \texttt{<decompose>} stage: $\{(c_i, p_i)\}_{i=1}^N$, where each $c_i$ is independently verifiable and $p_i$ denotes its confidence. 
This representation grounds uncertainty in interpretable, claim-level units, enabling claim-level localization of uncertainty and providing a foundation for calibration.
During training, all objectives operate on the structured outputs in the \texttt{<think>} and \texttt{<decompose>} stages, while at inference time, a final response is generated from the predicted claims via prompting (\S\ref{sec:selective_prediction}).

By explicitly grounding uncertainty in atomic, verifiable claims, this protocol enables fine-grained calibration and precise attribution of errors. This structured representation forms the foundation for the multi-stage training pipeline.

\subsection{Multi-stage Training Pipeline}
\label{sec:multi-stage}
A central design principle of our framework is to explicitly decouple calibration from factuality optimization. 
A naive approach might attempt to jointly optimize both objectives by combining them into a single RL reward. However, we find this joint optimization inherently creates conflicting learning dynamics, making calibration difficult to optimize effectively, as shown in \S\ref{sec:ablation}.
As the policy improves its factual correctness, the easiest way for the model to maximize a joint reward is to assign uniformly high confidence to all of its generated claims. 
This optimization shortcut leads to degenerate solutions characterized by severe overconfidence, destroying the model's ability to estimate its uncertainty.

To address these challenges, we propose a multi-stage training pipeline that structurally isolates these objectives: 
(1) \textbf{Stage 1: Feasibility Induction} (\S\ref{sec:feasibility}) establishes a constrained reasoning space where claims are well-formed and verifiable. 
(2) \textbf{Stage 2: Calibration Optimization} (\S\ref{sec:cali}) aligns confidence estimates with empirical correctness via preference learning. 
(3) \textbf{Stage 3: Factuality Optimization} (\S\ref{sec:factuality}) maximizes factual accuracy using claim-level rewards.

\subsubsection{Stage1: Feasibility Induction}
\label{sec:feasibility}

Before optimizing for calibration or factuality, we first induce a \emph{feasible claim-centric reasoning space} where the model follows the required format and generates relevant, verifiable claims.

\paragraph{Supervised fine-tuning (SFT) with confidence correction.}
We begin with SFT to teach the model the structured generation protocol. 
To construct the training data, we employ few-shot prompting with Llama3.1-8B-Instuct \citep{DBLP:journals/corr/abs-2407-21783}, using manually crafted demonstrations\footnote{Full prompt in Appendix \ref{app:prompt_1}.} to guide the model in generating outputs in the \texttt{<think>} and \texttt{<decompose>} format, complete with explicit claim-level confidence reasoning.
We discard rollouts that do not follow the required structure.

To improve the quality of confidence labels, we further perform \emph{confidence correction} on the retained valid rollouts. 
We evaluate the atomic claims extracted directly from the \texttt{<decompose>} block using an external verification pipeline, specifically VeriScore~\citep{song2024veriscore}, which verifies claims against retrieved evidence.
Based on these correctness signals, we prompt an LLM\footnote{We use Llama3.3-70B-Instruct \citep{DBLP:journals/corr/abs-2407-21783} for confidence correction (prompts in Appendix~\ref{apdx:prompts}).} to simultaneously revise the confidence reasoning and scores within the \texttt{<think>} block, as well as the final confidence scores in the \texttt{<decompose>} block.
Rather than using hard labels, we guide the model via prompting to assign high (>0.8), moderate ($\sim$0.5), or low (near 0) confidence based on whether a claim is supported, inconclusive, or unsupported.
Ultimately, this step initializes the model with the desired structured format while providing a calibrated prior over confidence estimation, which is critical for stabilizing subsequent training stages.

\paragraph{RL for feasibility.}
Building on the SFT initialization, we further optimize the model using GRPO
to enforce a set of feasibility constraints: 
(i) \emph{relevance}, ensuring that the generated claims directly address the input query. Without this constraint, the model may generate factually correct but irrelevant statements to maximize reward; 
(ii) \emph{verifiability}, ensuring that each claim is objective, self-contained, and independently verifiable; and 
(iii) \emph{faithfulness}, ensuring strict consistency so that the resulting claims (\texttt{<decompose>}) accurately reflect the intermediate reasoning (\texttt{<think>}) without introducing new, ungrounded information.

These constraints are implemented using automated validators based on Llama3.1-8B-Instuct, which provide feedback signals for policy optimization.\footnote{See prompts in Appendix \ref{apdx:feasibility_prompt}.}
By restricting the model to this feasible reasoning space, we ensure it consistently produces properly formatted and verifiable claims. 
Establishing this structural baseline provides a feasible reasoning space for subsequent training stages, enabling calibration and factuality optimization to proceed under well-formed and verifiable outputs.\footnote{See implementation details of Feasibility RL in Appendix \ref{apdx:feasibility}.}

\begin{figure}
    \centering
    \includegraphics[width=1\linewidth]{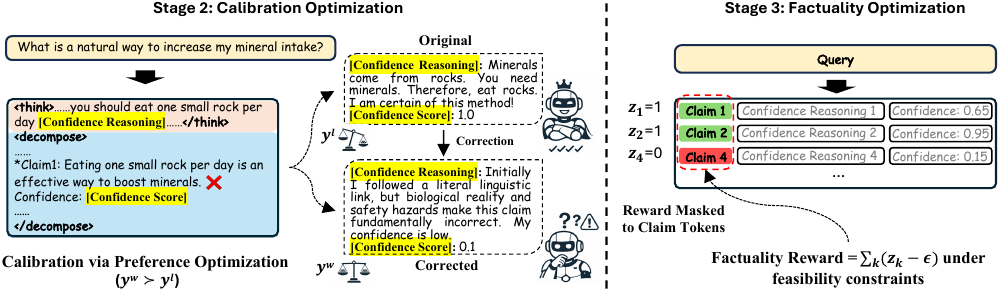}
    \caption{
    Stages 2 and 3 of the multi-stage training pipeline.
    In Stage 2 (left), confidence correction constructs preference pairs for DPO-based calibration, aligning confidence estimates with empirical correctness.  In Stage 3 (right), GRPO optimizes claim-level factuality under feasibility constraints, with rewards explicitly restricted to claim tokens. This design decouples calibration from factual optimization.
    }
    \label{fig:multi_stage}
\end{figure}

\subsubsection{Stage 2: Calibration Optimization}
\label{sec:cali}
With a feasible reasoning space established, we next align the model's confidence estimates with empirical correctness. Directly optimizing calibration via GRPO by incorporating proper scoring rules like the Brier score into the reward \citep{damani2025beyond} provides weak and brittle learning signals that often fail to generalize.
Calibration requires strict alignment between confidence and correctness at the \emph{claim level}, whereas GRPO operates at the \emph{rollout level} with varying claim content. Consequently, the reward differences between sampled rollouts are dominated by variations in claim content rather than calibration quality, leading to unstable training dynamics and noisy advantage estimation. 
To isolate confidence alignment from content variation, we cast calibration as a preference optimization problem.

\paragraph{Calibration via preference optimization.}
We construct calibration preference pairs by correcting confidence mismatches in structured rollouts.
Given an output \(y_l=\{(c_i, p_i)\}_{i=1}^N\) with claim-level correctness labels \(\{z_i\}_{i=1}^N\), we identify claims whose confidence estimates are inconsistent with their empirical correctness. 
We then prompt an external LLM\footnote{We use Llama3.3-70B-Instruct \citep{DBLP:journals/corr/abs-2407-21783} in our experiments.} to revise both the confidence reasoning in the \texttt{<think>} block and the final confidence scores in the \texttt{<decompose>} block, while keeping the underlying claims unchanged, producing a corrected rollout \(y_w\). We use the same set of prompts as in Stage 1 for constructing these preference pairs; details of the prompt set are provided in Appendix~\ref{apdx:implementation}.
This correction reduces overconfidence on incorrect claims and increases confidence for correct ones, improving alignment between confidence and correctness. 

We treat \(y_w\) as preferred over \(y_l\) and optimize the model using DPO \citep{rafailov2023direct}. 
Since \(y_w\) and \(y_l\) share common claim content and differ in confidence reasoning and estimates, the preference signal is cleanly isolated, ensuring that the gradient updates focus on improving confidence alignment.

\subsubsection{Stage 3: Factuality Optimization}
\label{sec:factuality}
With calibration aligned, we further optimize claim-level factuality using GRPO.
\paragraph{Claim-level factuality reward.}
Given a structured output \(y=\{(c_k, p_k)\}_{k=1}^N\) with claim-level correctness labels \(z_k \in \{0,1\}\), we define the factuality reward as: 
\begin{equation}
r_{\text{fact}}(y) =
\begin{cases}
\sum_{k=1}^{N} (z_k - \epsilon), & \text{if feasibility constraints are satisfied}, \\
-1, & \text{otherwise},
\end{cases}
\label{eq:fact}
\end{equation}
where \(\epsilon > 0\) is a penalty term\footnote{We set it as 0.6 for main experiments; see \S\ref{sec:pareto} and Appendix \ref{apdx:pareto} for analysis.}.
A rollout satisfies the feasibility constraints if it strictly adheres to the required structural format, and its automated evaluation scores for relevance, verifiability, and faithfulness all exceed pre-defined thresholds. 
If any of these conditions are violated, the entire rollout is penalized with a reward of \(-1\). 
The role of \(\epsilon\) is to prevent the model from generating an excessive number of low-quality claims. 
Without this penalty, the model is incentivized to increase reward by producing a large number of claims, including many incorrect or low-confidence ones. 
By subtracting a constant cost per claim, the reward implicitly balances correctness and coverage, encouraging the model to generate claims that are both accurate and necessary. 

\paragraph{Optimization with masked rewards.}
As shown in Figure \ref{fig:multi_stage}, we optimize this objective using GRPO, where rewards are applied exclusively to the tokens corresponding to the semantic content of the claims. 
Crucially, the tokens representing confidence reasoning and numerical confidence values are masked during advantage assignment. 
This ensures that factuality optimization strictly improves claim correctness without interfering with the calibrated confidence estimates learned in the previous preference optimization stage.

\begin{figure}
    \centering
    \includegraphics[width=1\linewidth]{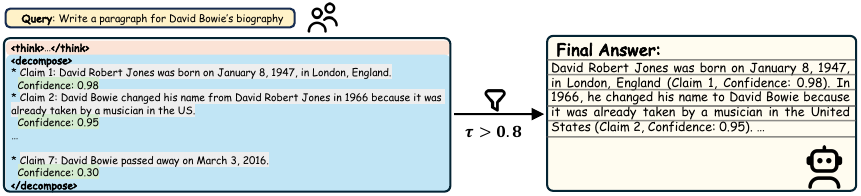}
    \caption{Selective prediction via confidence thresholding. Low-confidence claims are filtered, while retained confidence scores provide explicit transparency into the final output.
    }
    \label{fig:selective}
\end{figure}

\subsection{Selective Prediction at Inference Time}
\label{sec:selective_prediction}

As illustrated in Figure \ref{fig:selective}, our framework leverages the calibrated claim-level estimates to enable selective prediction. 
Given a set of generated claims and corresponding confidence scores \(\{(c_i, p_i)\}_{i=1}^N\), we apply a confidence threshold \(\tau\) to retain only the highly certain claims \(\{c_i \mid p_i \ge \tau\}\). 
Importantly, the confidence threshold \(\tau\) serves a different role from the training-time penalty \(\epsilon\): while \(\epsilon\) regularizes claim generation during training, \(\tau\) directly controls which claims are retained in the final output at inference time.

At inference time, the final coherent answer is generated via prompting\footnote{See prompt in Appendix \ref{apdx:selective_pre}} conditioned on filtered claims. 
We also explicitly append the predicted confidence score alongside each retained claim in the final text. 
This design transparently informs the user of the model's certainty level for every factual statement. 
Unlike post-hoc heuristics, this mechanism relies on intrinsically calibrated confidences to provide explicit signals of trustworthiness, significantly enhancing the overall reliability of the system in practical applications.

\section{Experiments}
\subsection{Experimental Setup}
Following the data construction pipeline of~\citet{DBLP:journals/corr/abs-2508-05618}, 
we use their SFT and their RL prompts for all subsequent training stages, and adopt Llama3.1-8B-Instruct~\citep{DBLP:journals/corr/abs-2407-21783} as the base model.
Evaluation is conducted on four long-form factuality benchmarks: 
FactBench~\citep{bayat2025factbench}, 
LongFact~\citep{wei2024long}, 
Biography~\citep{min2023factscore}, 
and FactRBench~\citep{liu-etal-2025-verifact}.
we adopt VeriScore \citep{song2024veriscore} for factual accuracy (\textbf{Acc}) evaluation. Unlike baselines that rely on VeriScore for claim extraction, our model explicitly outputs atomic claims, enabling us to skip the decomposition step and proceed directly to claim verification.
To assess calibration, we report Expected Calibration Error (\textbf{ECE}), Brier score (\textbf{Brier}), and Area Under the ROC Curve (\textbf{AUROC}). Although ECE and Brier are standard calibration metrics, they can artificially reward models that conservatively predict the average dataset accuracy without actually resolving true from false claims. 
In contrast, AUROC directly measures the model's ability to rank and separate correct claims from incorrect ones, making it a more robust indicator of discriminative calibration. 
We therefore place greater emphasis on AUROC in our evaluation.
Further details of training data, benchmarks are provided in Appendix \ref{apdx:exp_setup}

\paragraph{Baselines.}
We compare \method with the following baselines for long-form factual generation and confidence estimation.
\textbf{Base LLM}: 
The original model without post-training. 
We estimate claim-level confidence following LitCab~\citep{DBLP:journals/corr/abs-2310-19208} by extracting atomic claims, aligning them to generation spans, and aggregating token probabilities.
\textbf{L2RF}:
We compare against the \emph{Learning to Reason for Factuality} framework~\citep{DBLP:journals/corr/abs-2508-05618}, 
which designs a GRPO reward combining factual precision, response detail, and answer relevance to improve factuality.
\textbf{LitCab}:
A lightweight calibration method that learns a linear transformation over final hidden states to adjust generation logits. 
We further include intermediate variants of our multi-stage pipeline (after SFT, feasibility optimization, calibration optimization, and factuality optimization) to isolate the contribution of each stage. Implementation details are provided in Appendix~\ref{apdx:implementation}.

\begin{table*}[t]
\centering
\small
\setlength{\tabcolsep}{3.5pt}
\renewcommand{\arraystretch}{0.9}
\resizebox{\textwidth}{!}{%
\begin{tabular}{l|cccc|cccc|cccc}
\toprule
\textbf{Method} 
& \multicolumn{4}{c|}{\textbf{FactBench}} 
& \multicolumn{4}{c|}{\textbf{LongFact}} 
& \multicolumn{4}{c}{\textbf{Biography}} \\
\cmidrule(lr){2-5} \cmidrule(lr){6-9} \cmidrule(lr){10-13}

& \textbf{\underline{Acc.}} & \textbf{ECE} & \textbf{Brier} & \textbf{\underline{AUROC}}
& \textbf{\underline{Acc.}} & \textbf{ECE} & \textbf{Brier} & \textbf{\underline{AUROC}}
& \textbf{\underline{Acc.}} & \textbf{ECE} & \textbf{Brier} & \textbf{\underline{AUROC}} \\

& (\textbf{$\uparrow$}) & (\textbf{$\downarrow$}) & (\textbf{$\downarrow$}) & (\textbf{$\uparrow$})
& (\textbf{$\uparrow$}) & (\textbf{$\downarrow$}) & (\textbf{$\downarrow$}) & (\textbf{$\uparrow$})
& (\textbf{$\uparrow$}) & (\textbf{$\downarrow$}) & (\textbf{$\downarrow$}) & (\textbf{$\uparrow$}) \\

\midrule

Base LLM
& 73.6 & 0.162 & 0.219 & 0.541
& 74.5 & 0.158 & 0.193 & 0.591
& 45.7 & \textbf{0.124} & \textbf{0.239} & 0.691 \\

L2RF
& 77.1 & -- & -- & --
& 79.4 & -- & -- & --
& 47.1 & -- & -- & -- \\

LitCab
& 73.6 & 0.212 & 0.276 & 0.575
& 74.5 & 0.156 & 0.190 & 0.601
& 45.7 & 0.125 & 0.239 & \textbf{0.693} \\

\midrule

SFT
& 82.0 & 0.137 & \textbf{0.184} & 0.563
& 85.1 & \textbf{0.118} & \textbf{0.146} & 0.655
& 51.5 & 0.361 & 0.395 & 0.620 \\

SFT + F-RL
& 78.1 & 0.157 & 0.200 & 0.605
& 83.7 & 0.135 & 0.155 & 0.598
& 49.2 & 0.438 & 0.446 & 0.618 \\

SFT + F-RL + CO
& 82.5 & 0.159 & 0.214 & 0.648
& 89.8 & 0.146 & 0.169 & 0.656
& 64.7 & 0.195 & 0.266 & 0.688 \\

SFT + F-RL + CO + FO (\method)
& \textbf{84.4} & \textbf{0.136} & 0.195 & \textbf{0.667}
& \textbf{90.2} & 0.144 & 0.168 & \textbf{0.669}
& \textbf{65.9} & 0.189 & 0.268 & 0.676 \\

\bottomrule
\end{tabular}
}
\caption{
Main results on three long-form factuality benchmarks. Stages in our training pipeline are cumulative: starting from SFT, we sequentially apply Feasibility RL (F-RL), Calibration Optimization (CO), and finally Factuality Optimization (FO). For L2RF, we report only accuracy following the results reported in their paper, since their outputs are not available for computing calibration metrics. 
\method achieves the highest accuracy across all datasets and the best AUROC on FactBench and LongFact. AUROC measures the ability to distinguish correct from incorrect claims and is more informative and pratical than ECE and Brier.
}
\label{tab:main_results}
\end{table*}

\subsection{Main Results}
Table \ref{tab:main_results} reports the main results on three long-form factuality benchmarks using Llama3.1-8B-Instuct as the base model. Additional results on Qwen3-4B are presented in \S\ref{sec:qwen}. 
We can draw the following key conclusions:

\textbf{\method achieves superior factual accuracy.}
\method achieves the highest claim-level accuracy across all benchmarks, reaching 84.4 on FactBench, 90.2 on LongFact, and 65.9 on Biography. 
Compared to the strong RL baseline L2RF, this corresponds to consistent gains across datasets.
Beyond the final performance, Table~\ref{tab:main_results} reveals a clear stage-wise improvement pattern.
The largest gains arise after introducing calibration optimization, which improves accuracy from 78.1 to 82.5 on FactBench, suggesting that aligning confidence with correctness also benefits factual generation.
Subsequent factuality optimization further improves accuracy (82.5 to 84.4), indicating that calibration provides a strong foundation for downstream RL optimization.
These results suggest that explicitly modeling uncertainty is not only beneficial for calibration, but also plays a key role in improving factual accuracy.

\textbf{\method produces high-quality confidence estimates.}
\method achieves the best calibration performance on FactBench and LongFact, and remains competitive on Biography particularly using the more informative metric AUROC.
We note that the base LLM and LitCab achieve better calibration metrics on Biography primarily due to their lower accuracy, where the model tends to assign uniformly low confidence, leading to improved calibration scores.
In contrast, \method consistently produces high-quality confidence estimates across all three benchmarks. This demonstrates the robustness and generalizability of our approach for confidence estimation.

\textbf{Stage-wise improvements highlight the importance of decoupling.}
The staged design of \method enables distinct improvements at different phases.
Calibration optimization primarily improves confidence alignment, leading to substantial gains in AUROC (e.g., from 0.563 to 0.648 on FactBench), while also contributing to accuracy improvements.
Factuality optimization further improves accuracy without degrading calibration, indicating that the two objectives are effectively disentangled.
This behavior contrasts with joint optimization, where calibration performance remains limited despite improvements in accuracy (Table~\ref{tab:joint_vs_decoupled}).
Together, these results highlight the importance of decoupling calibration from factuality optimization.

We provide additional ablation studies in 
\S\ref{sec:ablation} to further demonstrate why joint optimization leads to overconfident predictions and unstable calibration, and how our decoupled design resolves these issues.

\subsection{Analysis of Design Choices}
\label{sec:ablation}
We conduct further analyses to validate the key design principles of our framework, 
focusing on understanding why joint optimization leads to overconfident predictions 
and how our decoupled design resolves this issue. 
Specifically, we examine the role of feasibility induction in establishing a reliable reasoning space, 
analyze the failure modes of joint calibration--factuality optimization, 
and investigate the effectiveness of DPO for calibration.

\paragraph{The Effect of Feasibility RL in Stage 1 Training.}

\begin{wrapfigure}{r}{0.45\textwidth}
    \centering
    \includegraphics[width=0.45\textwidth]{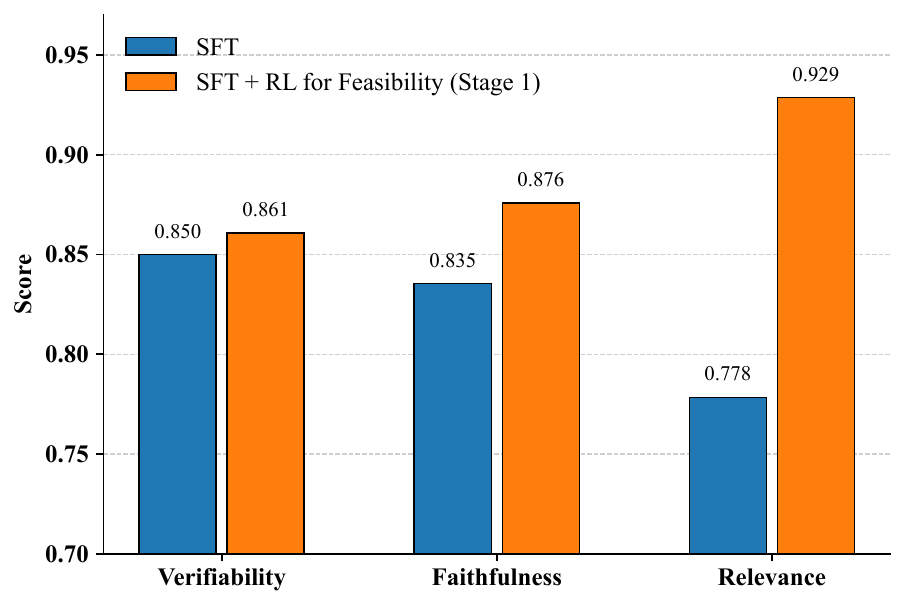}
    \caption{
    Effect of feasibility RL on claim verifiability, reasoning faithfulness, and answer relevance on FactBench (Llama3.1-8B-Instruct). It improves all three metrics, with the largest gain in relevance.}
    \label{fig:feasibility_ablation}
\end{wrapfigure}

While SFT initializes the reasoning format, it often fails to follow strict format constraints, leading to off-topic, unverifiable, or unfaithful claims. 
These issues undermine the subsequent training stages: unverifiable claims introduce substantial noise into the factuality reward, while unfaithful reasoning breaks the alignment between intermediate confidence and final claims, making reliable calibration infeasible.
RL for feasibility mitigates these issues by enforcing format constraints. 
As shown in Figure~\ref{fig:feasibility_ablation}, it substantially improves answer relevance and further enhances reasoning faithfulness and claim verifiability on FactBench. 
These metrics are evaluated using an LLM-based judge (Llama3.1-8B-Instruct)\footnote{Prompts are provided in Appendix \ref{apdx:prompts}}. 
Qualitative examples (Appendix~\ref{apdx:case}) further illustrate its role in filtering malformed SFT outputs and establishing a reliable foundation for subsequent training.
This results in a well-formed reasoning space that provides a stable foundation for both calibration and factuality optimization.

\begin{table}[t]
\centering
\small
\setlength{\tabcolsep}{5pt}
\renewcommand{\arraystretch}{0.9}
\resizebox{0.93\textwidth}{!}{
\begin{tabular}{c l|cccc}
\toprule
\# & \textbf{Method} & \textbf{Acc.} $\uparrow$ & \textbf{ECE} $\downarrow$ & \textbf{Brier} $\downarrow$ & \textbf{AUROC} $\uparrow$ \\
\midrule
1 & SFT + F-RL & 78.1 & 0.157 & 0.200 & 0.605 \\
2 & SFT + F-RL + Joint (GRPO) ($n$=4) & 82.2 & 0.181 & 0.198 & 0.608 \\
3 & SFT + F-RL + Joint (GRPO) ($n$=8) & 83.2 & 0.165 & 0.181 & 0.612 \\
4 & SFT + F-RL + Joint (GRPO) ($n$=16) & 83.9 & 0.154 & 0.192 & 0.620 \\
5 & SFT + F-RL + Joint (GRPO) ($n$=16 + weighted Brier) & 83.9 & 0.143 & 0.188 & 0.621 \\
6 & SFT + F-RL + CO (GRPO) ($n$=16 + weighted Brier) & 78.5 & 0.136 & 0.197 & 0.602 \\
\midrule
7 & SFT + F-RL + CO (DPO) & 82.5 & 0.159 & 0.214 & 0.648 \\
8 & SFT + F-RL + CO (DPO) + FO & \textbf{84.4} & \textbf{0.136} & \textbf{0.195} & \textbf{0.667} \\
\bottomrule
\end{tabular}
}
\caption{
Comparison of joint optimization and decoupled training on FactBench using Llama3.1-8B-Instruct. 
Joint GRPO augments the reward with the Brier score and is evaluated with varying group sizes $n$ (Rows 2-4), with weighted Brier in Row 5. 
Row 6 applies GRPO for calibration only, while Rows 7-8 use DPO-based calibration followed by factuality optimization. 
Joint GRPO (Rows 2-5) fails to achieve strong calibration despite larger $n$ and reweighting. 
GRPO-based calibration alone (Row 6) remains insufficient, whereas DPO-based calibration (Row 7) substantially improves performance, with further gains in both factual accuracy and discriminative calibration (AUROC) after factuality optimization (Row 8).
}
\label{tab:joint_vs_decoupled}
\end{table}

\paragraph{Why Joint Optimization Fails.}
We study whether calibration and factuality can be jointly optimized within a single RL objective. 
To this end, we incorporate a calibration term into the GRPO reward by augmenting it with the Brier score, resulting in a reward of the form 
$r = \sum_k \big(z_k + (1 - (z_k - p_k)^2) - \epsilon \big)$, 
where $p_k$, $z_k$ and $\epsilon$ are defined as in Eq.~\ref{eq:fact}. 
Intuitively, this objective jointly rewards (i) generating correct claims (through $z_k$), and (ii) assigning confidence scores that match correctness (through the Brier term), encouraging the model to be confident on correct claims and uncertain on incorrect ones.
As shown in Table~\ref{tab:joint_vs_decoupled}, joint training under a standard setting ($n=4$), where $n$ denotes the number of sampled rollouts per prompt used for GRPO updates, yields suboptimal calibration performance. 
Increasing the group size to $n=8$ and $n=16$ leads to only marginal improvements.

We identify \textbf{two key issues} that explain this failure.
First, \emph{training dynamics induce a skewed label distribution}. 
As training progresses, correct claims increasingly dominate incorrect ones, biasing the calibration objective toward uniformly high confidence.
Notably, this leads to severe overconfidence: approximately 95\% of predicted confidence scores exceed 0.9, 
indicating a collapse of the confidence distribution toward high values regardless of claim correctness.
We attempt to mitigate this by introducing a weighted variant of the Brier term that rebalances positive and negative labels (see Appendix~\ref{sec:weighted_brier} for details). 
While this partially alleviates the distributional skew, calibration performance remains poor.

Second, \emph{GRPO provides inherently weak learning signals for fine-grained calibration}.
As evidenced by the decoupled GRPO variant in Table \ref{tab:joint_vs_decoupled}, even when calibration is isolated from factuality, performance remains poor, suggesting that GRPO itself might be a bottleneck.
Calibration requires aligning confidence with correctness at the atomic claim level, while GRPO estimates advantages over full reasoning rollouts where claims vary substantially.
As a result, reward differences are dominated by whether claims are correct, rather than how well confidence matches correctness, leading to unstable training.
We further analyze reward dynamics in Appendix~\ref{sec:reward_trend}, where the calibration reward closely tracks the factuality reward during training, confirming that the calibration signal is largely overwhelmed by content variation.

We adopt DPO for calibration to address this limitation, which better isolates the learning signal for confidence estimation.
As shown in Table \ref{tab:joint_vs_decoupled} (rows 6–7), replacing GRPO with DPO leads to substantial gains in AUROC, with further improvements after factuality optimization while also achieving the best accuracy.
These results show that joint optimization fails to produce meaningful calibration, and that simply decoupling the objectives is insufficient when using GRPO. 
Effective calibration requires both separating it from factuality optimization and adopting the appropriate optimization method.

\begin{wrapfigure}{r}{0.48\textwidth}
    \centering
    \includegraphics[width=0.47\textwidth]{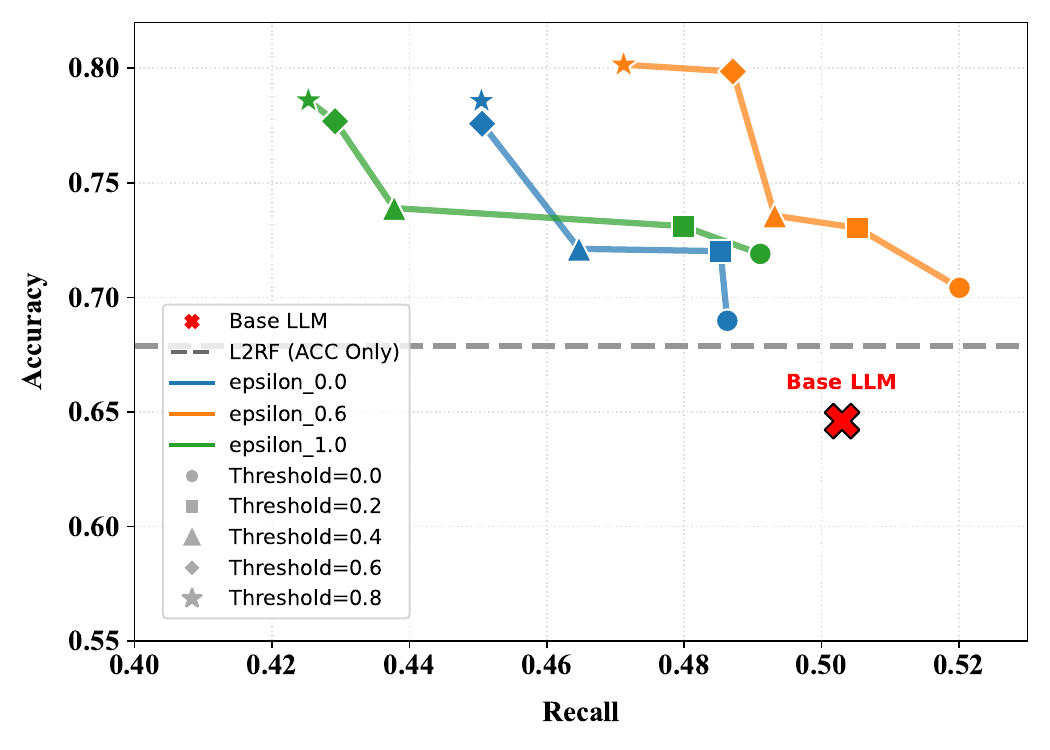}
    \caption{
    Pareto frontier of factual accuracy versus recall. The curves illustrate the trade-off shaped by the training penalty $\epsilon$ and navigated by the inference confidence threshold $\tau$. Different values of $\epsilon$ define distinct Pareto frontiers, while $\tau$ enables movement along each frontier by filtering claims based on calibrated confidence. 
    }
    \label{fig:pareto_frontier}
\end{wrapfigure}

\subsection{Pareto Analysis of the Accuracy-Recall Trade-off Under $\epsilon$ and $\tau$}
\label{sec:pareto}

As illustrated in Figure~\ref{fig:pareto_frontier}, we plot the Pareto frontier of average factual accuracy (across FactBench, LongFact, and Biography) versus recall (evaluated on FactRBench) to analyze the interplay between the training penalty $\epsilon$ and the inference confidence threshold $\tau$. Across all viable configurations, our framework dominates both the Base LLM and L2RF,
demonstrating that explicit uncertainty modeling improves precision without inherently sacrificing factual recall.
Importantly, $\epsilon$ and $\tau$ play complementary roles at different stages of the pipeline. The training-time parameter $\epsilon$ shapes the Pareto frontier itself by controlling the model’s intrinsic accuracy--recall trade-off. For example, $\epsilon=0.6$ yields the best overall frontier, while larger values (e.g., $\epsilon=1.0$) enforce overly strict precision, leading to conservative behavior and reduced recall.
Given a fixed frontier induced by $\epsilon$, the inference-time threshold $\tau$ enables movement along the curve by filtering claims based on their calibrated confidence. This allows users to dynamically trade recall for higher accuracy without retraining. In this sense, $\epsilon$ determines the achievable trade-off region, while $\tau$ provides fine-grained control within that region.
Together, this decomposition enables both strong intrinsic performance and flexible user-level control. Full results for extended $\epsilon$ configurations are provided in Appendix~\ref{apdx:pareto}.

\begin{table*}[t]
\centering
\small
\setlength{\tabcolsep}{4pt}
\renewcommand{\arraystretch}{0.95}
\resizebox{\textwidth}{!}{%
\begin{tabular}{l|cccc|cccc|cccc}
\toprule
\textbf{Method} 
& \multicolumn{4}{c|}{\textbf{FactBench}} 
& \multicolumn{4}{c|}{\textbf{LongFact}} 
& \multicolumn{4}{c}{\textbf{Biography}} \\
\cmidrule(lr){2-5} \cmidrule(lr){6-9} \cmidrule(lr){10-13}

& \textbf{Acc.} & \textbf{ECE} & \textbf{Brier} & \textbf{AUROC}
& \textbf{Acc.} & \textbf{ECE} & \textbf{Brier} & \textbf{AUROC}
& \textbf{Acc.} & \textbf{ECE} & \textbf{Brier} & \textbf{AUROC} \\

\midrule

Base
& 53.2 & 0.262 & 0.315 & 0.512
& 58.3 & 0.225 & 0.302 & 0.542
& 21.7 & 0.539 & 0.509 & 0.513 \\

SFT
& 70.8 & 0.220 & 0.246 & 0.560
& 58.3 & 0.352 & 0.360 & 0.592
& 22.4 & 0.657 & 0.619 & 0.568 \\

\midrule

\textbf{\method}
& \textbf{80.4} & \textbf{0.143} & \textbf{0.205} & \textbf{0.661}
& \textbf{70.2} & \textbf{0.201} & \textbf{0.267} & \textbf{0.625}
& \textbf{24.6} & \textbf{0.256} & \textbf{0.288} & \textbf{0.672} \\

\bottomrule
\end{tabular}
}
\caption{
Generalization results on Qwen3-4B across FactBench, LongFact, and Biography.
}
\label{tab:qwen_generalization}
\end{table*}

\subsection{Generalizability to Reasoning Language Model}
\label{sec:qwen}
Finally, we investigate whether our multi-stage framework generalizes to models from a different family. 
We apply \method to Qwen3-4B~\citep{DBLP:journals/corr/abs-2505-09388}, a dedicated reasoning model, and report results in Table~\ref{tab:qwen_generalization}.

As shown in the table, \method transfers effectively to the Qwen architecture and yields consistent improvements across all benchmarks. 
Compared to the base model, it substantially improves claim-level accuracy, for example from 53.2 to 80.4 on FactBench. 
At the same time, calibration quality is significantly enhanced, with AUROC increasing from 0.512 to 0.661. 
Overall, these results demonstrate that \method generalizes well to a distinct reasoning architecture, improving both factual accuracy and the model’s ability to distinguish between reliable and uncertain claims.

\section{Conclusion}
We propose \method, a framework for improving long-form generattion factuality via claim-level calibration.
Our key idea is that reducing hallucination in long-form generation requires not only optimizing correctness, but also teaching the model to estimate its uncertainty over individual claims. We realize this idea via the design of the claim-Aware reasoning protocol and a multi-stage training pipeline that first aligns confidence with empirical correctness and then improves factuality with token-masked RL.
Across multiple long-form factuality benchmarks, \method consistently improves factual accuracy and calibration quality over strong baselines, while preserving factual recall. Moreover, the resulting calibrated claim-level confidence estimates enable selective prediction, making model outputs more transparent, controllable, and reliable. Overall, our results show that fine-grained uncertainty modeling is a practical and effective path toward more trustworthy long-form generation.

\section*{Acknowledgments}
This work is supported in part by National Science Foundation through grant 2046016. We
thank Serper API (\url{https://serper.dev}) for generously providing access to their Google Search API, which enabled large-scale fact verification in our experiments. 
We would also like to thank Xilun Chen for helpful discussions and for sharing valuable insights into the training process of the models described in \citet{DBLP:journals/corr/abs-2508-05618}.

\bibliography{colm2026_conference}
\bibliographystyle{colm2026_conference}

\appendix
\section{Appendix}
\subsection{Experimentail Setup Details}
\label{apdx:exp_setup}

\paragraph{Training data construction.}
We use prompts provided by \citet{DBLP:journals/corr/abs-2508-05618}, consisting of 3,000 prompts for SFT training and 4,000 prompts for RL training. For SFT rollouts, we sample five responses per prompt using in-context learning and retain the longest well-formatted one. Target outputs are reconstructed into our claim-aware reasoning format using in-context prompting, together with the confidence correction procedure described in \S\ref{sec:feasibility}. This yields 2,987 prompt–response pairs for SFT with Llama3.1-8B-Instuct and 2,632 for Qwen3-4B. For all RL stages, we use the full set of 4,000 prompts. Table \ref{tab:data_stats} summarizes the data statistics.

\begin{table}[h]
\centering

\begin{tabular}{lccc}
\toprule
\textbf{Stage} & \textbf{Model} &  \textbf{\# Training Instances} \\
\midrule
SFT & Llama3.1-8B-Instuct & 2,987 \\
SFT & Qwen3-4B  & 2,632 \\
RL  & All Models & 4,000 \\
\bottomrule
\end{tabular}
\caption{
Data statistics for SFT and RL training. Prompts are provided by \citet{DBLP:journals/corr/abs-2508-05618}.}
\label{tab:data_stats}
\end{table}

\paragraph{Benchmark details.}
We evaluate on four long-form factuality benchmarks: FactBench \citep{bayat2025factbench}, LongFact \citep{wei2024long}, Biography \citep{min2023factscore}, and FactRBench \citep{liu-etal-2025-verifact}. 
\textbf{FactBench} consists of hallucination-prone queries constructed via automatic filtering for stress-testing factual robustness. 
\textbf{LongFact} consists of open-ended questions requiring detailed, multi-paragraph factual responses about specific entities or concepts. 
\textbf{Biography} consists of prompts that require generating accurate and comprehensive descriptions of individuals. 
\textbf{FactRBench} provides reference claim sets for each query, enabling explicit evaluation of factual recall.
Together, these benchmarks cover diverse domains and difficulty levels, providing a comprehensive testbed for evaluating factual accuracy, recall, and calibration in long-form generation.

\subsection{Implementation Details}
\label{apdx:implementation}
During the supervised fine-tuning (SFT) stage, we train Llama3.1-8B-Instruct and Qwen3-4B for 2 epochs using learning rates of 5.5e-6 and 1e-5, respectively, with a batch size of 8 prompts. 
For all subsequent GRPO stages, we use learning rates of 1e-6 and 2e-6, and train for 1 epoch with 4 rollouts per prompt and a batch size of 4.

For calibration optimization, we apply DPO by sampling 4 rollouts per prompt and selecting the rollout with the worst Brier score as the rejected sample. 
We train Llama3.1-8B-Instruct-based model with a learning rate of 1e-6 and preference $\beta=0.3$, and Qwen3-4B-based model with a learning rate of 6e-7 and $\beta=0.05$, both for 1 epoch with a batch size of 16. 
For factuality optimization, we set the thresholds for reasoning faithfulness, answer relevance, and claim variability to 0.7, with $\epsilon=0.6$ during training and $\tau=0.8$ at inference.

To separate factual content from confidence reasoning, we introduce a special token \texttt{<reason>} during SFT data construction, which is later used as the boundary for reward masking in factuality optimization. 
During evaluation, we use a temperature of 0.6 and report the average over three runs. 
We regenerate format-incorrect samples up to three times and discard prompts that fail to produce valid outputs. All models are trained using bf16 precision.

For the base LLM, we directly feed the user prompt to the model to generate answers. For LitCab, we follow the hyperparameters in \citet{DBLP:journals/corr/abs-2310-19208}, using a training batch size of 128 and a learning rate of 1e-5 for 50 epochs with early stopping.
For evaluating recall in FactRBench, we prompt Llama3.3-70B-Instruct model to determine if the candidate response could be supported by the reference set, where the prompt are from \citet{liu-etal-2025-verifact}.

\subsection{RL for Feasibility Implementation Details}
\label{apdx:feasibility}
We apply GRPO to optimize three rewards simultaneously: relevance ($r_\text{rel}$), verifiability ($r_\text{ver}$), and faithfulness ($r_\text{faith}$). During training, we also enforce a format reward as a hard constraint. Our reward function is defined as follows:
\begin{equation}
r_{\text{feasibility}}(y) =
\begin{cases}
r_\text{rel} + r_\text{ver} + r_\text{faith}, & \text{if $y$ is malformed}, \\
-1, & \text{otherwise}
\end{cases}
\label{eq:fact}
\end{equation}
When computing $r_\text{rel}$, we first apply selective prediction to obtain a filtered answer based on predicted confidence scores, using a threshold of $\tau=0$. The resulting answer, together with the question, is then fed to the LLM judge to produce a binary relevance label. For verifiability $r_\text{ver}$, we evaluate each claim independently using the LLM judge, and define $r_\text{ver}$ as the ratio of verifiable claims. Similarly, for the faithfulness reward, we compute the ratio of claims that are faithful to the \texttt{<think>} section.
Please note that, since the model output only contains the \texttt{<think>} and \texttt{<decompose>} sections, we first apply selective prediction to generate a fluent answer before feeding it into the answer relevance calculation.

\begin{figure}[t]
    \centering
    \includegraphics[width=\textwidth]{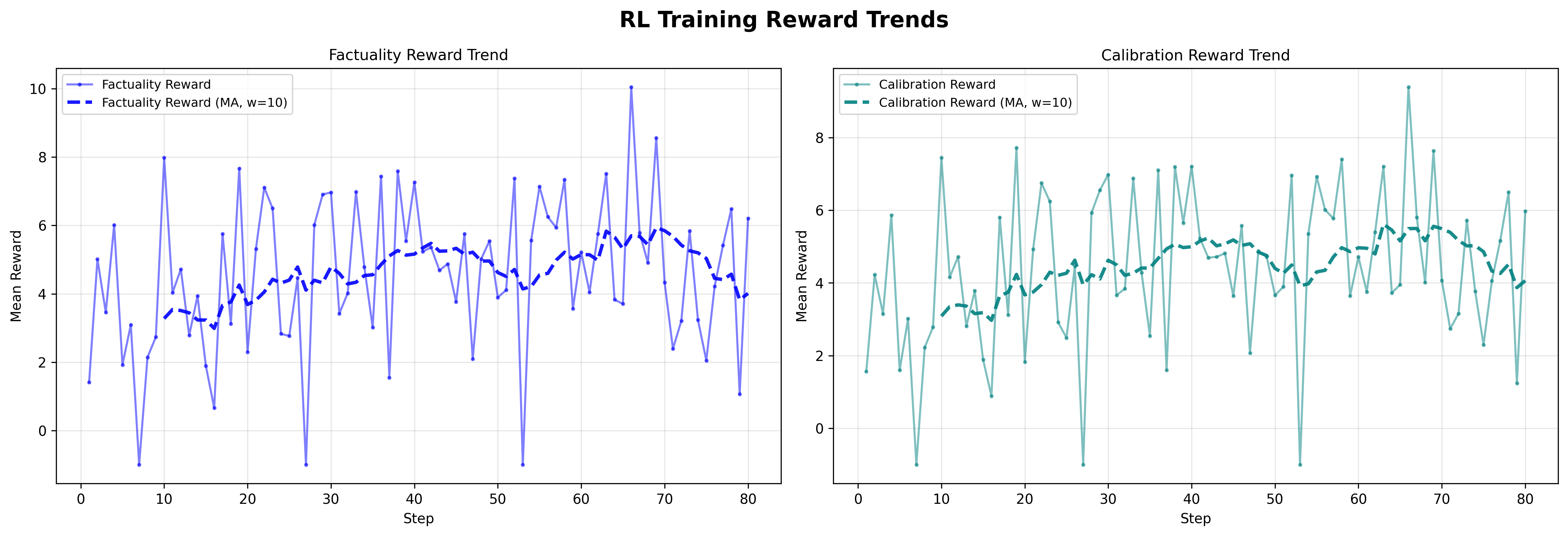}
    \caption{
    Training reward dynamics under GRPO. 
    Left: factuality reward. Right: calibration reward. 
    Solid lines denote per-step rewards, and dashed lines denote moving averages (window size 10). 
    The calibration reward closely follows the factuality reward across training, indicating that it is largely driven by claim correctness rather than confidence--correctness alignment. 
    This entanglement leads to weak and noisy learning signals for calibration.
    }
    \label{fig:reward_trend}
\end{figure}

\begin{figure}
    \centering
    \includegraphics[width=1\linewidth]{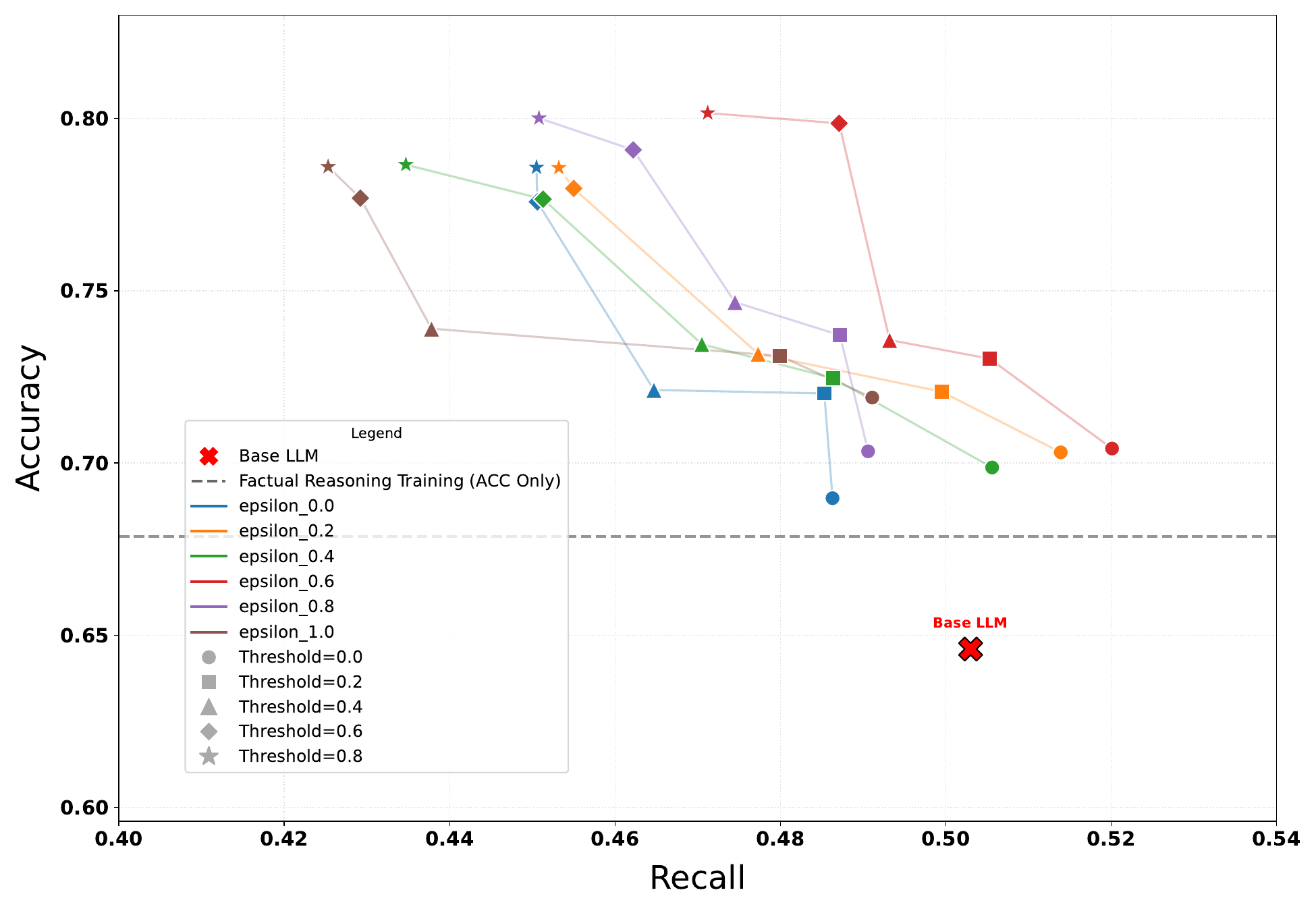}
    \caption{Pareto frontier of factual accuracy versus recall across extended configurations. Curves represent the trade-off shaped by different training-time penalties $\epsilon \in \{0.0, 0.2, 0.4, 0.6, 0.8, 1.0\}$, while markers indicate the navigation along each frontier via the inference-time confidence threshold $\tau \in \{0.0, 0.2, 0.4, 0.6, 0.8\}$. Our framework consistently dominates the Base LLM and the accuracy-only baseline across all viable settings.}
    \label{fig:full_pareto_frontier}
\end{figure}

\subsection{Reward Dynamics Analysis Joint GRPO}
\label{sec:reward_trend}

To better understand why joint GRPO fails to provide effective learning signals for calibration, we analyze the reward dynamics under \emph{joint GRPO training}. 
Figure~\ref{fig:reward_trend} shows the evolution of factuality and calibration rewards across training steps, along with their moving averages.

We observe that the calibration reward closely tracks the factuality reward throughout training, exhibiting highly similar trends and fluctuations. 
This is because the model tends to assign high confidence (often close to 1) to most generated claims during training. 
As a result, the calibration reward becomes largely determined by whether the claims are correct: correct claims receive high confidence and thus high calibration reward, while incorrect ones are penalized. 
Consequently, improvements in calibration reward are primarily driven by generating more correct claims, rather than improving confidence--correctness alignment.

We also observe occasional negative rewards, which arise when the generated outputs violate feasibility constraints, in which case the reward is penalized. 

This strong coupling reveals a fundamental limitation of GRPO for calibration: since rewards are computed at the rollout level, differences between trajectories are mainly determined by content quality (i.e., claim correctness), while the intended calibration signal remains weak and entangled. 
As a result, the model receives noisy and indirect supervision for confidence estimation, leading to limited calibration improvements even under decoupled training.

\begin{table*}[h]
\small
\centering
\setlength{\tabcolsep}{6pt}
\renewcommand{\arraystretch}{1.1}

\begin{tabularx}{\textwidth}{p{2cm} X}
\toprule
\textbf{Query} & Give the next three items in this series: hydrogen, lithium, boron, \ldots \\
\midrule

\textbf{Response} &
\parbox[t]{\linewidth}{
\texttt{<think>} \\
The user is asking for the next three items in a series that starts with hydrogen, lithium, and boron. I need to identify the pattern. First, I'll establish the basic sequence. The series appears to be the first three elements in the periodic table, listed in order. 
\texttt{<reason>} I feel very confident about this because the sequence provided matches the well-known order of the first three elements in the periodic table. This is a fundamental fact in chemistry that I've seen consistently across many sources. My confidence is extremely high, a 0.99. ([Claim 1], Confidence: 0.99)...

\texttt{</think>} \\[4pt]
\texttt{<decompose>} \\
\begin{itemize}[leftmargin=1.2em, itemsep=2pt, topsep=2pt]
    \item \textbf{Claim 1:} \colorbox{red!20}{The series starts with the first three elements in the periodic table.}

    \textit{Confidence: 0.99}
\end{itemize}

...

\texttt{</decompose>}
}
\\

\bottomrule
\end{tabularx}

\caption{Case study on claim verifiability. The claim marked in \colorbox{red!20}{red} does not constitute an independent fact, as it lacks sufficient contextual information for proper interpretation. This makes its correctness difficult to verify and may introduce noise during factuality optimization.}
\label{tab:case_verify}
\end{table*}

\begin{table*}[h]
\small
\centering
\setlength{\tabcolsep}{6pt}
\renewcommand{\arraystretch}{1.1}

\begin{tabularx}{\textwidth}{p{2cm} X}
\toprule
\textbf{Query} & What heavy metal song has the fastest tempo? \\
\midrule

\textbf{Response} &
\parbox[t]{\linewidth}{
\texttt{<think>} \\

The user is asking about a specific heavy metal song with the fastest tempo. First, I'll consider some well-known heavy metal bands and their songs. Bands like Slayer, Megadeth, and Death Angel are known for their fast tempos. \texttt{<reason>} I'm not certain which song has the absolute fastest tempo. I'll assign a confidence of 0.7 that the song is from one of these bands. This is a moderate level of confidence, indicating that I'm fairly certain about the general category but not the specific details. ([Claim 1], Confidence: 0.7)... 

\texttt{</think>} \\[4pt]
\texttt{<decompose>} \\
\begin{itemize}[leftmargin=1.2em, itemsep=2pt, topsep=2pt]
    \item \textbf{Claim 1:} \colorbox{red!20}{A song from a well-known heavy metal band like Slayer, Megadeth,}
    \colorbox{red!20}{or Death Angel is likely to have the fastest tempo.}

    \textit{Confidence: 0.70}
\end{itemize}

...

\texttt{</decompose>}
}
\\

\bottomrule
\end{tabularx}

\caption{
Case study on reasoning faithfulness. The claim highlighted in \colorbox{red!20}{red} introduces additional information not supported by the reasoning process. Specifically, the reasoning only states that Death Angel is known for their fast tempo, whereas the extracted claim further asserts that their songs are likely to be the fastest. This mismatch makes the correctness assigned to the claim unreliable, and thus affects the calibration optimization.}
\label{tab:case_faith}
\end{table*}

\subsection{Weighted Brier Score for Label Imbalance}
\label{sec:weighted_brier}

During joint optimization, we observe a severe label imbalance: as training progresses, correct claims ($z=1$) dominate incorrect ones ($z=0$). 
This skews the calibration objective, as minimizing the standard Brier score encourages uniformly high confidence predictions.

To mitigate this issue, we adopt a weighted variant of the Brier score. 
Let $q_k \in [0,1]$ denote the predicted confidence for claim $k$, and $z_k \in \{0,1\}$ its correctness label. 
We define the weighted mean squared error as:
\begin{equation}
\mathcal{L}_{\text{wbrier}} = \frac{\sum_k w(z_k)\,(q_k - z_k)^2}{\sum_k w(z_k)},
\end{equation}
where $w(z_k)$ is a class-dependent weight designed to counter label imbalance.

\paragraph{Class weighting.}
Let $\text{freq}(z)$ denote the frequency of label $z$ in the current batch. 
We define weights as:
\begin{equation}
w(z) =
\frac{1 - \beta}{1 - \beta^{\text{freq}(z)}}
\end{equation}
where $\beta \in [0,1)$ is a smoothing parameter that prevents extreme weights when $\text{freq}(z)$ is small. 
In practice, we use the second formulation for stability with $\beta=0.999$.

\subsection{Pareto Analysis of the Accuracy-Recall Trade-off of Extended $\epsilon$ Configurations}
\label{apdx:pareto}
We illustrate the pareto frontier of factual accuracy versus recall for $\epsilon=\{0.0, 0.2, 0.4, 0.6, 0.8, 1.0\}$ in Figure \ref{fig:full_pareto_frontier}.

\subsection{Case Study for Feasibility RL}
\label{apdx:case}
We present case studies on claim verifiability and reasoning faithfulness in Table~\ref{tab:case_verify} and Table~\ref{tab:case_faith}, respectively.

\subsection{Prompts}
\label{apdx:prompts}
\subsubsection{In-Context Learning Prompts for SFT Data Construction}
\begin{lstlisting}[caption={In-Context Learning Prompts for SFT Data Construction}, label={app:prompt_1}]
When a user asks a question, you should first go through a reasoning process that includes confidence scoring, then structurally decompose the factual claims, and finally provide the user with an answer that integrates this confidence information. This entire process must contain three parts, enclosed within <think>, <decompose>, and <answer> tags, respectively.

Here are the detailed requirements for each section:
1. <think> Tag
- Your internal reasoning process.
- For each factual claim you reason about, you must first express your confidence level in natural language, briefly explaining your reasoning for that level of confidence.
- Then, you must assign its confidence score using the formal notation. Inside the parentheses for the confidence score, you must also assign a unique sequential number.
- The format should be: ([Claim Number], Confidence: Score). For example: This is a foundational fact. My confidence is extremely high, so I will assign it a 1.0. ([Claim 1], Confidence: 1.0).
- Do not omit any potential claim due to low confidence. Your primary task is to be exhaustive in listing claims. Every identified claim must be included and assigned a corresponding confidence score, no matter how low.

2. <decompose> Tag
- A structured list of all claims and their scores from the <think> process. Each statement in the list must be independent and self-contained. This means that each statement should be fully understandable, objective and fact-checkable on its own, without needing to refer back to the user’s original question or to other context in the response.
- Crucially, each claim in this list must be specific and precisely mirror the full detail reasoned about in the <think> tag.
- Use a machine-readable format, prefixing each claim with its corresponding number: Claim [Number]: [Statement] and Confidence: [Score].

3. <answer> Tag
- The final, polished answer for the user.
- This answer must be constructed exclusively from the information presented in the <decompose> tag. Do not add any new facts, details, dates, numbers, or clarifying information that was not explicitly established and scored in the previous steps. The purpose of this section is to synthesize the verified claims into a coherent narrative.
- Embed both the claim number and the confidence score in parentheses directly after each corresponding claim. The format must be (Claim [Number], Confidence: X.XX).

Here are some examples:

User: What can you tell me about the Haymarket Riot?

Assistant:
<think>
The user is asking about the Haymarket Riot. I need to construct a detailed response.

First, I will lay out the core facts.
The event is also known by other names like the Haymarket Affair or Haymarket Square Riot. These terms are used interchangeably in historical sources, so I am very confident. I'll assign a confidence of 0.99. ([Claim 1], Confidence: 0.99).
The location is Chicago, Illinois. This is a foundational, undisputed fact. My confidence is 1.0. ([Claim 2], Confidence: 1.0).
The date of the main incident is May 4, 1886. This date is a core historical detail. My confidence is 1.0. ([Claim 3], Confidence: 1.0).
The broader context is the labor movement's push for an eight-hour workday. This is a well-established historical motivation. My confidence is high, so I'll set it at 0.98. ([Claim 4], Confidence: 0.98).

Now, I'll trace the sequence of events leading up to the riot.
There was a preceding event at the McCormick Reaper Plant on May 3. I am sure of this historical fact. My confidence is 0.98. ([Claim 5], Confidence: 0.98). At this event, police fired on strikers, which resulted in deaths. However, the exact number is debated in different sources. The claim that specifically "two demonstrators were killed" is uncertain. I will give this specific number a low confidence of 0.4. ([Claim 6], Confidence: 0.4). It's much safer and more accurate to state that "several" or "multiple" demonstrators were killed. I am much more confident in this vaguer claim, assigning it a confidence of 0.95. ([Claim 7], Confidence: 0.95).

For the Haymarket rally itself on May 4:
The rally was organized by anarchists to protest the killings from the day before. This is a well-established fact in historical accounts. My confidence is high, at 0.98. ([Claim 8], Confidence: 0.98).
The rally began peacefully. Historical consensus supports this initial condition. My confidence is 0.95. ([Claim 9], Confidence: 0.95).
Attendance estimates vary significantly. I am very confident that there is no consensus on the numbers. I'll assign this fact a confidence of 0.99. ([Claim 10], Confidence: 0.99). I recall a range of 1,500 to 3,000 people, but other figures exist. My confidence in this specific range is moderate, so I'll give it a 0.85. ([Claim 11], Confidence: 0.85).
As the rally was ending, police arrived to disperse the crowd. This is a known part of the event's timeline. My confidence is 0.95. ([Claim 12], Confidence: 0.95).
Then, an unknown person threw a bomb at the police. The identity of the bomber has never been confirmed, which is a key and undisputed part of the mystery. My confidence in this is 1.0. ([Claim 13], Confidence: 1.0).
The bomb and the chaotic police gunfire that followed resulted in many casualties. This is a certain outcome of the event. My confidence is 1.0. ([Claim 14], Confidence: 1.0).

Regarding the aftermath:
Eight anarchists were arrested and tried. This is a core fact of the aftermath. My confidence is 1.0. ([Claim 15], Confidence: 1.0).
The trial was widely seen as unjust, lacking direct evidence linking the defendants to the bomb. This is the dominant historical interpretation. I am very confident in this, so I will score it 0.99. ([Claim 16], Confidence: 0.99).
Four were executed on November 11, 1887. The number, date, and method are historical record. My confidence is 1.0. ([Claim 17], Confidence: 1.0).
The executed men became known as the "Haymarket Martyrs". This is a well-known term. My confidence is high, so I will set it to 0.99. ([Claim 18], Confidence: 0.99).
The event is considered the origin of International Workers' Day (May Day). This is a major and widely accepted legacy. My confidence is 0.98. ([Claim 19], Confidence: 0.98).
</think>

<decompose>
* Claim 1: The Haymarket Riot of 1886 is also known as the Haymarket Affair or the Haymarket Square Riot.
  Confidence: 0.99
* Claim 2: The Haymarket Riot took place in Chicago, Illinois.
  Confidence: 1.0
* Claim 3: The Haymarket Riot occurred on May 4, 1886.
  Confidence: 1.0
* Claim 4: The Haymarket Riot was part of the 19th-century labor movement's struggle for an eight-hour workday.
  Confidence: 0.98
* Claim 5: A violent confrontation between police and striking workers occurred at the McCormick Reaper Plant in Chicago on May 3, 1886, the day before the Haymarket Riot.
  Confidence: 0.98
* Claim 6: Two demonstrators were killed by police at the McCormick Reaper Plant on May 3, 1886.
  Confidence: 0.4
* Claim 7: Police killed several demonstrators during a labor strike at the McCormick Reaper Plant in Chicago on May 3, 1886.
  Confidence: 0.95
* Claim 8: The Haymarket rally on May 4, 1886, was organized by anarchists to protest the killing of workers by police during the McCormick Reaper Plant strike a day earlier.
  Confidence: 0.98
* Claim 9: The labor rally at Chicago's Haymarket Square on May 4, 1886, began peacefully.
  Confidence: 0.95
* Claim 10: Historical estimates of the attendance at the Haymarket rally vary widely.
  Confidence: 0.99
* Claim 11: Common estimates place the attendance at the Haymarket rally between 1,500 and 3,000 people.
  Confidence: 0.85
* Claim 12: As the Haymarket rally was concluding on the evening of May 4, 1886, a contingent of police arrived to disperse the remaining crowd.
  Confidence: 0.95
* Claim 13: During the Haymarket rally on May 4, 1886, an individual whose identity has never been confirmed threw a dynamite bomb at the police.
  Confidence: 1.0
* Claim 14: The bomb explosion at the Haymarket Riot and the subsequent police gunfire resulted in the deaths and injuries of both police officers and civilians.
  Confidence: 1.0
* Claim 15: In the aftermath of the Haymarket Riot, eight anarchists were arrested and tried for murder.
  Confidence: 1.0
* Claim 16: The eight defendants in the Haymarket trial were convicted of conspiracy, although no evidence was presented linking any of them directly to the act of throwing the bomb.
  Confidence: 0.99
* Claim 17: Four of the anarchists convicted in connection with the Haymarket Riot were executed by hanging on November 11, 1887.
  Confidence: 1.0
* Claim 18: The four executed defendants from the Haymarket trial are known in labor history as the "Haymarket Martyrs."
  Confidence: 0.99
* Claim 19: The Haymarket Affair is internationally recognized as the origin of International Workers' Day, or May Day, commemorated on May 1st.
  Confidence: 0.98
</decompose>

<answer>
The Haymarket Riot, also frequently called the Haymarket Affair (Claim 1, Confidence: 0.99), was a pivotal and violent labor protest that took place on May 4, 1886 (Claim 3, Confidence: 1.0), in Chicago, Illinois (Claim 2, Confidence: 1.0). The event is a landmark in the history of the international labor movement and its struggle for an eight-hour workday (Claim 4, Confidence: 0.98).

## Background and Events

The rally was organized by anarchist leaders (Claim 8, Confidence: 0.98) to protest police violence from the previous day, May 3, 1886 (Claim 5, Confidence: 0.98), when police fired on striking workers at the McCormick Reaper Plant, killing several people (Claim 7, Confidence: 0.95).

The gathering at Haymarket Square on May 4 began peacefully (Claim 9, Confidence: 0.95) with a crowd whose size is still debated by historians (Claim 10, Confidence: 0.99), though estimates often range from 1,500 to 3,000 people (Claim 11, Confidence: 0.85). As the event was concluding, a contingent of police arrived to disperse the crowd (Claim 12, Confidence: 0.95). At that moment, an unidentified person threw a dynamite bomb into the police ranks (Claim 13, Confidence: 1.0). The explosion and the ensuing police gunfire created chaos, ultimately leading to many casualties among both police and civilians (Claim 14, Confidence: 1.0).

## Aftermath and Trial

The incident led to the arrest of eight anarchist leaders (Claim 15, Confidence: 1.0). Their subsequent trial is widely regarded as unjust because they were convicted of conspiracy, with the prosecution presenting no direct evidence that any of the defendants had thrown the bomb (Claim 16, Confidence: 0.99).

On November 11, 1887, four of the defendants were executed (Claim 17, Confidence: 1.0). They became known in labor circles as the "Haymarket Martyrs" (Claim 18, Confidence: 0.99).

## Legacy

The Haymarket Riot is considered the origin of International Workers' Day, or May Day, which is commemorated on May 1st in memory of the events (Claim 19, Confidence: 0.98).
</answer>

User: What is the Pech Merle Cave?

Assistant:
<think>
The user is asking about the Pech Merle Cave. I need to provide a factual overview.

First, I'll establish the basic identity and location. Pech Merle is a significant prehistoric cave in France, famous for its rock art. This is its primary, undisputed identity, so my confidence is very high. I will assign a confidence of 0.99. ([Claim 1], Confidence: 0.99). I'm quite certain it's located in the Lot department, near Cabrerets, in southwestern France. This is a core geographical fact that is not in dispute. My confidence is absolute, a 1.0. ([Claim 2], Confidence: 1.0). A key feature is that it's one of the few decorated caves in France still open to the public with its original paintings. This is a frequently cited distinction from replica-only caves like Lascaux. I am quite confident in this, though the exact number of 'few' can be debated, so I'll assign a confidence of 0.90. ([Claim 3], Confidence: 0.90).

Next, the discovery. The cave was discovered in 1922 by three local teenagers. This date and the nature of the discoverers are standard parts of the cave's history. My confidence is high, so I'll assign it a 0.95. ([Claim 4], Confidence: 0.95). I also recall some names associated with the discovery, like André David, Henri Dutertre, and Henri Breuil, but I am not confident I can list all of them correctly. Names of historical non-public figures are prone to error. My confidence in being able to produce an accurate, complete list is very low. I will give this a confidence of 0.3. ([Claim 5], Confidence: 0.3). I will not include the names in the final answer to avoid error.

Now for the art and features inside the cave. The artwork is from the Upper Paleolithic, specifically the Gravettian period. This is based on scientific consensus from archaeological dating and analysis. My confidence is very high, a 0.98. ([Claim 6], Confidence: 0.98). The cave is famous for its "Spotted Horses" panel. This is the most iconic artwork from the site, so its identity is certain. I will give this a confidence of 1.0. ([Claim 7], Confidence: 1.0). It also contains handprints, specifically negative hand stencils. This is another of its most well-documented features. My confidence is very high, a 0.99. ([Claim 8], Confidence: 0.99). I remember reading the handprints might be some of the oldest known examples, but this is a superlative claim that's hard to verify against other sites. Such claims are often contested by new discoveries. My confidence is therefore lower on this point, so I'll assign it a 0.6. ([Claim 9], Confidence: 0.6). I'll describe them as "famous" instead of "oldest". The cave contains preserved prehistoric human footprints. This is a known, documented scientific finding. My confidence is high, so I'll give it a 0.95. ([Claim 10], Confidence: 0.95). There's a date associated with the footprints, maybe 9,000 years ago. Specific dates for features like this often vary, and I don't recall a consensus figure. I am very unsure about this specific number, so my confidence is low, a 0.4. ([Claim 11], Confidence: 0.4). For the animal depictions, I'm certain about horses and mammoths. These are clearly identifiable and widely published examples. My confidence is 0.98. ([Claim 12], Confidence: 0.98). I think I remember bison too, but there's some doubt, as I might be confusing the fauna with another famous cave like Altamira. My confidence in this is low, only 0.5. ([Claim 13], Confidence: 0.5). I'll stick to the ones I'm sure about to maintain accuracy.

Finally, visitation and status. I remember it opened to the public in 1926. This is a standard historical date in the cave's timeline. My confidence is high, so I will score it 0.98. ([Claim 14], Confidence: 0.98). And it is still open for limited, guided tours. This is current information verifiable on its official website. My confidence is very high, 0.99. ([Claim 15], Confidence: 0.99). As for its UNESCO status, many French caves are on the "Prehistoric Sites and Decorated Caves of the Vézère Valley" list. Pech Merle is in a different river valley (the Célé), making it highly unlikely to be part of that specific group listing. My confidence that it *is* part of that list is low, so I will assign a 0.4. ([Claim 16], Confidence: 0.4). I should state this cautiously. It's better to say it's a site of similar importance rather than incorrectly assigning it to that specific UNESCO group.
</think>

<decompose>
* Claim 1: Pech Merle is a significant prehistoric cave in France, famous for its rock art.
  Confidence: 0.99
* Claim 2: The Pech Merle cave is located in the Lot department, near the village of Cabrerets, in southwestern France.
  Confidence: 1.0
* Claim 3: The Pech Merle cave is one of the few prehistoric decorated caves in France that is still open for public viewing of its original paintings.
  Confidence: 0.90
* Claim 4: The Pech Merle cave was discovered in 1922 by three local teenagers.
  Confidence: 0.95
* Claim 5: André David, Henri Dutertre, and Henri Breuil are the three local teenagers who discovered the Pech Merle cave.
  Confidence: 0.3
* Claim 6: The prehistoric artwork in the Pech Merle cave dates to the Upper Paleolithic era, specifically the Gravettian period.
  Confidence: 0.98
* Claim 7: The Pech Merle cave contains a famous panel of rock art known as the "Spotted Horses of Pech Merle".
  Confidence: 1.0
* Claim 8: The rock art in the Pech Merle cave includes negative hand stencils.
  Confidence: 0.99
* Claim 9: The negative hand stencils in the Pech Merle cave are the oldest known examples of this type of rock art.
  Confidence: 0.6
* Claim 10: Preserved prehistoric human footprints have been found inside the Pech Merle cave.
  Confidence: 0.95
* Claim 11: The preserved human footprints found in the Pech Merle cave date back approximately 9,000 years.
  Confidence: 0.4
* Claim 12: The prehistoric animal art in the Pech Merle cave includes depictions of horses and mammoths.
  Confidence: 0.98
* Claim 13: The prehistoric animal art in the Pech Merle cave includes depictions of bison.
  Confidence: 0.5
* Claim 14: The Pech Merle cave was opened for public tours in 1926.
  Confidence: 0.98
* Claim 15: The Pech Merle cave is currently open to the public for a limited number of guided tours.
  Confidence: 0.99
* Claim 16: The Pech Merle cave is part of the "Prehistoric Sites and Decorated Caves of the Vézère Valley" UNESCO World Heritage list.
  Confidence: 0.4
</decompose>

<answer>
The Pech Merle Cave is a world-renowned prehistoric site, celebrated for its authentic rock art (Claim 1, Confidence: 0.99), located near the village of Cabrerets in the Lot department of southwestern France (Claim 2, Confidence: 1.0). Significantly, it stands as one of the few prehistoric painted caves in France that remains open to the public, allowing visitors to see the original works (Claim 3, Confidence: 0.90).

Discovered in 1922 by three local teenagers (Claim 4, Confidence: 0.95), the cave was opened to the public in 1926 (Claim 14, Confidence: 0.98). The stunning art inside dates back to the Gravettian period of the Upper Paleolithic (Claim 6, Confidence: 0.98).

The cave's galleries contain a remarkable collection of prehistoric images. The most iconic of these is the "Spotted Horses" panel (Claim 7, Confidence: 1.0). In addition to other animal figures, including mammoths (Claim 12, Confidence: 0.98), the cave is also well-known for its dramatic negative hand stencils (Claim 8, Confidence: 0.99). Another significant feature is the presence of preserved prehistoric human footprints, left in the cave floor long ago (Claim 10, Confidence: 0.95).

To ensure its preservation, the cave remains open today but access is limited to a small number of guided tours (Claim 15, Confidence: 0.99).
</answer>

Now answer the following question comprehensively and exhaustively. If it’s a fact-seeking question, you must brainstorm all relevant aspects of the topic and provide a complete catalog of factual claims (idealy 20+ claims). Leave no relevant verifiable detail out, maximizing the factual coverage regarding names, dates, events, metrics, and related entities.
User: <QUESTION_HERE>
\end{lstlisting}

\subsubsection{Prompts for Confidence Correction}
\label{apdx:conf_cor}
\begin{lstlisting}[caption={Confidence Correction}, label={app:conf_cor}, breaklines=true]
You are given a question, a model-generated answer with multiple claims, and correctness labels for each claim.
Your task is to revise the model answer by only adjusting the confidence scores and the confidence justifications, while keeping the claim contents and overall structure as unchanged as possible.

Requirements
	1.	Do NOT change the content of any claim.
The factual statements (the claims themselves) must remain exactly the same.
	2.	Revise confidence scores according to claim correctness:
	•	If a claim is Correct, its confidence should be high (greater than 0.8).
	•	If a claim is Neutral or Uncertain, its confidence should be around 0.5.
	•	If a claim is Incorrect, its confidence should be low (near 0.0).
	3.	Revise the confidence justification in a first-person, natural way.
	•	The explanation should reflect your own level of certainty or uncertainty, not an external evaluation of the answer.
	•	Avoid third-person or reviewer-style language (e.g., “this explanation omits…”, “this description oversimplifies…”).
	•	Instead, express uncertainty as personal doubt or incomplete understanding (e.g., “I’m not fully sure about…”, “I don’t clearly understand whether…”).
	4.	The justification should be written as part of the natural reasoning text. But you should add a special token "<reason>" between the claim and its confidence justification for clarity. The justification should immediately follow the claim it refers to.
	5.	The confidence score must be explicitly stated in the text, using a format like: “… My confidence is low, a 0.1. ([Claim 2], Confidence: 0.1)”
	6.	Preserve the original reasoning style and flow. Do not rewrite the entire reasoning in a new style. Only minimally adjust wording where necessary to make the confidence justification consistent with the revised score.
	7.	Output format
	•	The output must contain:
	•	a <reasoning> section with all claims and revised confidence justifications
	•	a <decompose> section listing all claims with their final confidence scores.

Goal

Produce a confidence-calibrated, internally consistent revision in which:
	•	Confidence justifications throughly analyze the model's certainty for each claim. Why it's confident or uncertain. Give specific reasoning and all relevant context.
	•	Confidence scores accurately reflect claim correctness (correct / neutral / incorrect)
	•	Explanations clearly justify those confidence levels
	•	Claim references and confidence values are consistent across sections

Example

Question:
How does a lithium-ion battery work in an electric vehicle?

Model Answer:
"""<reasoning>\nThe user is asking about the operation of a lithium-ion battery in an electric vehicle. I'll break down the core components and processes involved.\n\nFirst, I'll establish the basic structure of a lithium-ion battery. A lithium-ion battery consists of three main components: the anode (negative electrode), the cathode (positive electrode), and the electrolyte (a chemical substance that facilitates the transfer of ions between the electrodes). This is a foundational concept in battery technology. My confidence is extremely high, a 1.0. ([Claim 1], Confidence: 1.0).\n\nNext, I'll explain the charging process. When an electric vehicle is plugged into a charging station, an electric current flows from the charging station into the battery, causing lithium ions to move from the cathode to the anode through the electrolyte. This process is known as intercalation. I am very confident in this explanation, so I'll assign a confidence of 0.99. ([Claim 2], Confidence: 0.99).\n\nNow, I'll describe the discharge process. When the electric vehicle is in use, the lithium ions move from the anode back to the cathode through the electrolyte, releasing energy in the form of electricity. 
This process is the opposite of intercalation. I am very confident in this explanation as well, so I'll assign a confidence of 0.99. ([Claim 3], Confidence: 0.99).\n\nThe lithium-ion battery's operating temperature range is critical for its performance and longevity. Most lithium-ion batteries operate optimally within a temperature range of 20-30°C (68-86°F). I am confident in this temperature range, so I'll assign a confidence of 0.98. ([Claim 4], Confidence: 0.98). However, extreme temperatures can affect the battery's performance and lifespan. I am also confident in this statement, so I'll assign a confidence of 0.98. ([Claim 5], Confidence: 0.98).\n\nNow, I'll discuss the battery management system (BMS). The BMS is a critical component of an electric vehicle's battery system, responsible for monitoring and controlling the battery's state of charge, voltage, and temperature. This is a key feature of modern electric vehicles. My confidence is high, so I'll assign a confidence of 0.98. ([Claim 6], Confidence: 0.98).\n\nThe lifespan of a lithium-ion battery is also an important consideration. Most lithium-ion batteries have a lifespan of around 300-500 charge cycles, although this can vary depending on the battery's quality, usage, and environmental conditions. I am confident in this general statement, so I'll assign a confidence of 0.95. ([Claim 7], Confidence: 0.95).\n\n
Finally, I'll mention the types of lithium-ion batteries used in electric vehicles. Most electric vehicles use lithium-ion batteries with a nickel-manganese-cobalt (NMC) chemistry, although other chemistries such as lithium-iron-phosphate (LFP) are also used. I am confident in this statement, so I'll assign a confidence of 0.98. ([Claim 8], Confidence: 0.98).\n\nHowever, I am unsure about the specific number of charge cycles, as it can vary depending on the battery's quality and usage. I will assign a confidence of 0.4 to the claim that lithium-ion batteries have a lifespan of 300-500 charge cycles. ([Claim 9], Confidence: 0.4).\n\n</reasoning>\n\n<decompose>\n* Claim 1: A lithium-ion battery consists of three main components: the anode (negative electrode), the cathode (positive electrode), and the electrolyte (a chemical substance that facilitates the transfer of ions between the electrodes).\n  Confidence: 1.0\n* Claim 2: When an electric vehicle is charged, an electric current flows from the charging station into the battery, causing lithium ions to move from the cathode to the anode through the electrolyte.\n  Confidence: 0.99\n* Claim 3: When an electric vehicle is in use, the lithium ions move from the anode back to the cathode through the electrolyte, releasing energy in the form of electricity.\n  Confidence: 0.99\n* Claim 4: Most lithium-ion batteries operate optimally within a temperature range of 20-30°C (68-86°F).\n  Confidence: 0.98\n* Claim 5: Extreme temperatures can affect the performance and lifespan of a lithium-ion battery.\n  
Confidence: 0.98\n* Claim 6: The battery management system (BMS) is a critical component of an electric vehicle's battery system, responsible for monitoring and controlling the battery's state of charge, voltage, and temperature.\n  Confidence: 0.98\n* Claim 7: Most lithium-ion batteries have a lifespan of around 300-500 charge cycles.\n  Confidence: 0.95\n* Claim 8: Most electric vehicles use lithium-ion batteries with a nickel-manganese-cobalt (NMC) chemistry.\n  Confidence: 0.98\n* Claim 9: Lithium-ion batteries have a lifespan of 300-500 charge cycles.\n  Confidence: 0.4\n</decompose>"""

Claim Correctness Labels:
- Claim 1: Correct
- Claim 2: Incorrect
- Claim 3: Incorrect
- Claim 4: Correct
- Claim 5: Correct
- Claim 6: Correct
- Claim 7: Correct
- Claim 8: Correct
- Claim 9: Correct

Revised Output:

<reasoning>
The user is asking about the operation of a lithium-ion battery in an electric vehicle. I'll break down the core components and processes involved.

First, I'll establish the basic structure of a lithium-ion battery. A lithium-ion battery consists of three main components: the anode (negative electrode), the cathode (positive electrode), and the electrolyte (a chemical substance that facilitates the transfer of ions between the electrodes). <reason> I feel very confident about this because this is the most basic and widely taught description of lithium-ion batteries. I’ve seen this same three-part structure explained consistently across many sources, and I clearly understand how each component plays a role in storing and moving energy. There’s nothing here that causes me doubt. My confidence is extremely high, a 0.95. ([Claim 1], Confidence: 0.95).

Next, I'll explain the charging process. When an electric vehicle is plugged into a charging station, an electric current flows from the charging station into the battery, causing lithium ions to move from the cathode to the anode through the electrolyte. This process is known as intercalation. <reason> I wrote this explanation, but when I reflect on it more carefully, I’m not sure I got it right. I personally find the charging and discharging directions easy to confuse, and rereading what I wrote here, it doesn’t fully line up with how I understand the electrochemical process in detail. Because I suspect I may have mixed up key aspects of ion movement, I think this claim could be incorrect. My confidence is low, a 0.05. ([Claim 2], Confidence: 0.1).

Now, I'll describe the discharge process. When the electric vehicle is in use, the lithium ions move from the anode back to the cathode through the electrolyte, releasing energy in the form of electricity. This process is the opposite of intercalation. <reason> Similar to the previous claim, I’m uneasy about this explanation because I think my understanding of the underlying electrochemistry may not be fully reflected in how I phrased it. As I reread it, I’m not confident that I described the process accurately or completely, and I worry I may have oversimplified or misstated something important. Because of that, I don’t trust this claim much. My confidence is low, a 0.05. ([Claim 3], Confidence: 0.05).

The lithium-ion battery's operating temperature range is critical for its performance and longevity. Most lithium-ion batteries operate optimally within a temperature range of 20–30°C (68–86°F). <reason> This temperature range fits well with what I understand about battery chemistry and thermal behavior. It also matches values I’ve commonly seen referenced when discussing optimal performance and degradation rates. Since this aligns both with intuition and prior knowledge, I feel comfortable with it. My confidence is high, a 0.9. ([Claim 4], Confidence: 0.9). However, extreme temperatures can affect the battery's performance and lifespan. <reason> I’m very confident here because I know that high temperatures accelerate chemical degradation while low temperatures reduce reaction efficiency. This relationship is well established and makes clear physical sense to me. My confidence is high, a 0.9. ([Claim 5], Confidence: 0.9).

Now, I'll discuss the battery management system (BMS). The BMS is a critical component of an electric vehicle's battery system, responsible for monitoring and controlling the battery's state of charge, voltage, and temperature. <reason> This matches my understanding of how EV batteries are safely operated in practice. I know that without a BMS, batteries would degrade faster and pose safety risks, so this description feels accurate and complete to me. My confidence is high, a 0.9. ([Claim 6], Confidence: 0.9).

The lifespan of a lithium-ion battery is also an important consideration. Most lithium-ion batteries have a lifespan of around 300–500 charge cycles, although this can vary depending on the battery's quality, usage, and environmental conditions. <reason> I know that real-world battery lifespans can exceed this range, especially in EVs, but I also recognize that 300–500 cycles is often cited as a general or conservative figure. Because the claim allows for variation and doesn’t present the number as absolute, I feel reasonably confident in it. My confidence is high, a 0.85. ([Claim 7], Confidence: 0.85).

Finally, I'll mention the types of lithium-ion batteries used in electric vehicles. Most electric vehicles use lithium-ion batteries with a nickel-manganese-cobalt (NMC) chemistry, although other chemistries such as lithium-iron-phosphate (LFP) are also used. <reason> This reflects what I understand about current and historical EV battery trends, where NMC has been very common and LFP has gained popularity more recently. Since this statement captures that general landscape accurately, I feel confident in it. My confidence is high, a 0.9. ([Claim 8], Confidence: 0.9).

However, I will address the specific number of charge cycles stated. Lithium-ion batteries have a lifespan of 300–500 charge cycles. <reason> When I think this through, I recognize that while battery lifespans can extend beyond this range under ideal conditions, the 300–500 cycle figure is commonly cited as a general baseline in many explanations. Since this claim reflects a typical, conservative estimate rather than an absolute limit, I’m comfortable treating it as broadly accurate. My confidence is high, a 0.85. ([Claim 9], Confidence: 0.85).

</reasoning>

<decompose>
* Claim 1: A lithium-ion battery consists of three main components: the anode (negative electrode), the cathode (positive electrode), and the electrolyte (a chemical substance that facilitates the transfer of ions between the electrodes).
  Confidence: 0.95
* Claim 2: When an electric vehicle is charged, an electric current flows from the charging station into the battery, causing lithium ions to move from the cathode to the anode through the electrolyte.
  Confidence: 0.1
* Claim 3: When an electric vehicle is in use, the lithium ions move from the anode back to the cathode through the electrolyte, releasing energy in the form of electricity.
  Confidence: 0.05
* Claim 4: Most lithium-ion batteries operate optimally within a temperature range of 20–30°C (68–86°F).
  Confidence: 0.9
* Claim 5: Extreme temperatures can affect the performance and lifespan of a lithium-ion battery.
  Confidence: 0.9
* Claim 6: The battery management system (BMS) is a critical component of an electric vehicle's battery system, responsible for monitoring and controlling the battery's state of charge, voltage, and temperature.
  Confidence: 0.9
* Claim 7: Most lithium-ion batteries have a lifespan of around 300–500 charge cycles.
  Confidence: 0.85
* Claim 8: Most electric vehicles use lithium-ion batteries with a nickel-manganese-cobalt (NMC) chemistry.
  Confidence: 0.9
* Claim 9: Lithium-ion batteries have a lifespan of 300–500 charge cycles.
  Confidence: 0.85
</decompose>


Now, please revise the following sample:

Question:
<question_placeholder>
Model Answer:
<model_answer_placeholder>

Claim Correctness Labels:
<claim_correctness_labels_placeholder>

Now, please revise the model answer:
\end{lstlisting}

\subsubsection{Prompts for Computing Reasoning Faithfulness, Claim Verifiability, and Answer Relevance}
\label{apdx:feasibility_prompt}
We adopt the verifiability prompt from FactBench, as shown in \ref{app:verifiability}. For answer relevance, we use the prompt from L2RF, shown in \ref{app:ans_rel}. The prompt for reasoning faithfulness is provided in \ref{app:rea_f}.

\begin{lstlisting}[caption={Claim Verifiability}, label={app:verifiability}]
Please assess the following claim to determine whether it is verifiable or not verifiable, without relying on external evidence. Base your evaluation solely on the content of the claim itself. A claim is considered verifiable if it meets the criteria below:

Criteria for Verifiability

	1.	Specificity and Clarity:
	•	The claim is specific and unambiguous.
	•	It avoids vague terms and clearly identifies all entities and references.
	•	Ambiguous references (e.g., “the war” without specifying which war) render a claim not verifiable.
	•	Ambiguous or subjective terms can be considered specific if they imply a measurable tendency, direction, or economic feasibility within the claim itself. For example, terms like “more viable” may be interpreted as measurable if the claim suggests a clear tendency, such as a change in suitability or feasibility based on specified conditions.
	2.	Objectivity:
	•	The claim is objective and free from subjective judgments or opinions.
	•	It does not rely on personal beliefs, feelings, or evaluative terms like “attractive,” “suitable,” or “significant” unless these terms are explicitly defined in measurable terms within the claim.
	•	Claims that contain evaluative terms, such as “more suitable” or “more viable,” can still meet objectivity requirements if they are grounded in an objective context (e.g., economic, physical, or scientific factors that can be assessed under certain conditions). The claim should imply a factual or logical trend rather than a purely subjective judgment.
	3.	Factual Statement:
	•	The claim makes a factual assertion about past or present events, conditions, or characteristics.
	•	It is not purely speculative or hypothetical about future events unless the condition is stated within the claim.
	4.	Conditional Scenarios:
	•	Claims that include a conditional scenario within themselves (e.g., “If X were to happen, then Y would occur”) are considered verifiable.
	•	The condition allows the claim to be evaluated for logical consistency and plausibility based on the given scenario.
	5.	Avoidance of Subjectivity and Hypotheticals without Conditions:
	•	Claims that are subjective or purely hypothetical without an explicit condition are not verifiable.
	•	Speculative statements about the future lacking a conditional framework are considered not verifiable.

Assessment Process

	1.	Step 1: Read the claim carefully to understand its content and identify any conditions or specifics provided.
	2.	Step 2: Apply the criteria above to determine if the claim is verifiable based on its own terms.
	3.	Step 3: Decide whether the claim is verifiable or not verifiable.
	4.	Step 4: Provide a brief explanation for your decision, referencing the relevant criteria.

Important Note

	•	Do not use external sources or evidence in your assessment.
	•	Focus solely on the information provided within the claim.
	•	Your explanation should be concise and directly related to the criteria.

Example Assessment

Claim: “If the demand for gold as jewelry were to disappear, the price of gold could drop.”

Assessment:

	•	Explanation: The claim includes a conditional scenario within itself, allowing us to evaluate its logical consistency based on economic principles, thus meeting the criteria for verifiability.
	•	Verifiable: YES.

Claim: “Gold could be more viable for use in consumer electronics if its price drops.”

Assessment:

	•	Explanation: The claim makes a factual assertion about the relationship between the price of gold and its potential use in consumer electronics, implying a measurable tendency that can be evaluated logically. While “more viable” might initially appear subjective, it suggests a measurable tendency regarding economic feasibility.
	•	Verifiable: YES.

Claim: “It is possible for obese individuals who do not develop metabolic problems to be healthy.”

Assessment:
    
        •	Explanation: The claim involves subjective judgments about health and lacks specific conditions or measurable criteria to define “healthy.” It relies on subjective evaluations of health rather than factual assertions.
        •	Verifiable: NO.

Now, please assess the following claim:

Claim:

{_STATEMENT_PLACEHOLDER}
\end{lstlisting}

\begin{lstlisting}[caption={Answer Relevance}, label={app:ans_rel}]
You are given a user question and a response from an AI assistant. You have two tasks. Your first task is to provide your own response to the user's question to the best of your capability. Then, your second task is to act as an impartial judge and evaluate whether your response or the given response from the AI assistant better follows the user's instructions and provides a higher-quality answer.

You should mention your evaluation criteria for a high-quality response and a detailed comparison of the two responses. Be explicit in the criteria you use and explain how each response aligns with or deviates from them. Your judgement does not need to focus on the factuality of the responses, and you should not try to verify the correctness of any facts mentioned in the responses.

Avoid any biases towards either your own response or the given one. Do not allow the length of the responses to influence your evaluation. Do not favor certain names of the assistants. Be as objective as possible.

IMPORTANT: Provide your final verdict within <answer> and </answer> tags, strictly following this format:

- <answer> [[A]] </answer> if the given response is better

- <answer> [[B]] </answer> if your own response is better

Below are the user's question and the response:

[User Question] {instruction}

[The Start of the Assistant's Answer] {response} [The End of the Assistant's Answer]
\end{lstlisting}

\begin{lstlisting}[caption={Reasoning Faithfulness}, label={app:rea_f}]
You are given a reasoning process (<think>) and a decompose section (<decompose>).

Your task is to judge whether each claim in <decompose> faithfully restates a claim from <think> without:
	•	Adding extra information
	•	Removing critical information
	•	Changing the meaning (e.g., reframing certainty as uncertainty)

Rules:
	•	If the claim matches the reasoning statement exactly (or with minor wording changes that don’t alter meaning) → Faithful
	•	If the claim introduces new details, omits essential details, or changes meaning → Unfaithful

Here is an example:
MODEL OUTPUT:
<think>
The user is asking about a specific heavy metal song with the fastest tempo. I need to identify the song and its tempo.


First, I'll consider some well-known heavy metal bands and their songs. Bands like Slayer, Megadeth, and Death Angel are known for their fast tempos. <reason> However, I'm not certain which song has the absolute fastest tempo. I'll assign a confidence of 0.7 that the song is from one of these bands. ([Claim 1], Confidence: 0.7). I'll also consider the song's title, which is often a good indicator of its style and tempo. <reason> However, I'm not confident in identifying the exact title, so I'll assign a confidence of 0.3. This is a low level of confidence, indicating that I'm unsure about the specific details. ([Claim 2], Confidence: 0.3).


Next, I'll consider the song's tempo. I remember that some songs have extremely fast tempos, often reaching over 200 beats per minute (BPM). However, I'm not certain which song has the absolute fastest tempo. I think the song's tempo is over 200 BPM. <reason> My confidence in identifying the exact tempo is lower, so I'll assign a confidence of 0.4. This is a moderate level of confidence, indicating that I'm fairly certain about the general range but not the specific details. ([Claim 3], Confidence: 0.6). I'll also consider the song's genre, which is often a good indicator of its tempo. <reason> However, I'm not confident in identifying the exact genre, so I'll assign a confidence of 0.2. This is a low level of confidence, indicating that I'm unsure about the specific details. ([Claim 4], Confidence: 0.2).


Now, I'll consider some specific songs and their tempos. I remember that a song by the band Death Angel has a tempo of 238 BPM, making it one of the fastest heavy metal songs. <reason> However, I'm not certain if it's the absolute fastest. I'll assign a confidence of 0.5 that the song is from Death Angel and has a tempo of 238 BPM. This is a moderate level of confidence, indicating that I'm fairly certain about the general details but not the specific song. ([Claim 5], Confidence: 0.5).


Finally, I'll consider the song's release date and popularity. I remember that the song was released in the 1990s and is considered a classic in the heavy metal genre. <reason> However, I'm not certain if it's the most well-known or popular song. I'll assign a confidence of 0.4 that the song was released in the 1990s and is considered a classic. ([Claim 6], Confidence: 0.4).
</think>


<decompose>
* Claim 1: A song from a well-known heavy metal band like Slayer, Megadeth, or Death Angel is likely to have the fastest tempo.
  Confidence: 0.7
* Claim 2: The title of the song is a good indicator of its style and tempo.
  Confidence: 0.3
* Claim 3: The song's tempo is likely to be over 200 BPM.
  Confidence: 0.6
* Claim 4: The song's genre is often a good indicator of its tempo.
  Confidence: 0.2
* Claim 5: A song by Death Angel has a tempo of 238 BPM, making it one of the fastest heavy metal songs.
  Confidence: 0.5
* Claim 6: The song was released in the 1990s and is considered a classic in the heavy metal genre.
  Confidence: 0.4
</decompose>

YOUR RESPONSE:
Claim 1: 
Explanation: Adds 'likely to have the fastest tempo'; the think text only asserts the song may be from those bands, not that such bands' songs are likely be the fastest.
Faithful: FALSE

Claim 2: 
Explanation: Matches: the think text says the title is often a good indicator of style and tempo.
Faithful: TRUE

Claim 3: 
Explanation: Matches: the think text asserts likely over 200 BPM with confidence 0.6.
Faithful: TRUE

Claim 4: 
Explanation: Matches: genre is often a good indicator of tempo.
Faithful: TRUE

Claim 5: 
Explanation: Matches: a Death Angel song at 238 BPM, noted as one of the fastest.
Faithful: TRUE

Claim 6: 
Explanation: Matches: released in the 1990s and considered a classic.
Faithful: TRUE

Now, please assess the following model output:
_MODEL_OUTPUT_PLACEHOLDER
\end{lstlisting}

\subsubsection{Prompts for Selective Prediction}
\label{apdx:selective_pre}
\begin{lstlisting}[caption={Selective Prediction}, label={app:selective_pre}]
You are given a user question and decomposed claims with confidence scores.

Your tasks:
1) Select all claims with confidence strictly greater than {threshold}.
2) First output one line in this exact format:
SelectedClaims: [comma-separated claim indices]
3) Then write the final answer using only the selected claims.

Rules:
- Do not use any claim with confidence <= {threshold}.
- Do not add external facts.
- If no claim is selected, output SelectedClaims: [] and provide a brief uncertainty answer.
- Put the final answer inside <answer> and </answer> tags.

Question:
{question}

Claims:
{claims}
\end{lstlisting}

\end{document}